\documentclass[journal]{IEEEtran}
\newif\ifanonymized
\IEEEoverridecommandlockouts
\usepackage{dsfont}
\usepackage{amssymb}
\usepackage{amsmath,amsthm,amssymb,amsfonts}
\usepackage{ragged2e}
\usepackage{latexsym, amssymb, verbatim, amsmath}
\usepackage{amsmath,bm}
\usepackage{multirow}
\usepackage{mathtools}
\usepackage{subeqnarray}
\usepackage{cases}
\usepackage{amsmath}
\usepackage{amssymb}
\usepackage{mathrsfs}
\usepackage{graphicx}
\usepackage{subfigure}
\usepackage{booktabs}
\usepackage{color}
\usepackage{amsmath,amssymb,amsfonts}
\usepackage{algorithmic}
\usepackage{graphicx}
\usepackage{textcomp}
\usepackage{xcolor}
\usepackage{listings}
\usepackage[numbers,sort&compress]{natbib}
\usepackage[ruled,linesnumbered]{algorithm2e}

\usepackage{courier}
\usepackage{amssymb}
\usepackage{array}
\usepackage{hyperref}

\usepackage{mathtools}

\newtheorem{assumption}{Assumption}

\def\BibTeX{{\rm B\kern-.05em{\sc i\kern-.025em b}\kern-.08em
    T\kern-.1667em\lower.7ex\hbox{E}\kern-.125emX}}
\begin{document}

\title{FedOBP: Federated Optimal Brain Personalization through Cloud-Edge Element-wise Decoupling}

\ifanonymized
  \author{Anonymous Authors}
\else
\author{
    Xingyan Chen,~\IEEEmembership{Member,~IEEE,}
    Tian Du,
    Changqiao Xu,~\IEEEmembership{Senior Member,~IEEE,}
    Fuzhen Zhuang,~\IEEEmembership{Senior Member,~IEEE,} 
    Lujie Zhong,
    Gabriel-Miro Muntean,~\IEEEmembership{Fellow,~IEEE,}
    Enmao Diao,~\IEEEmembership{Member,~IEEE,}
    \IEEEcompsocitemizethanks{
        \IEEEcompsocthanksitem This work was supported in part by the National Natural Science Foundation of China under Grant 62302400, in part by the Fundamental Research Funds for the Beijing University of Posts and Telecommunications under Grant 2025AI4S08.
        \IEEEcompsocthanksitem X. Chen and C. Xu are with the Beijing Key Laboratory of MEMS Technology and Device Reliability for Industrial Internet, the School of Intelligent Engineering and Automation, Beijing University of Posts and Telecommunications, Beijing 100876, P. R. China.
        E-mail: \{chenxingyan94,cqxu\}@bupt.edu.cn.
        \IEEEcompsocthanksitem T. Du is with School of Business Administration, Southwestern University of Finance and Economics, Chengdu 611130, P. R. China.
        E-mail: dtian.nora@gmail.com.
        \IEEEcompsocthanksitem F. Zhuang is with the Institute of Artificial Intelligence and the School of Computer Science, Beihang University, Beijing 100191, China, and with the SKLSDE.
        E-mail: zhuangfuzhen@buaa.edu.cn.
        \IEEEcompsocthanksitem L. Zhong is with Information Engineering	College, Capital Normal University, Beijing 100048, P. R. China.
	    E-mail: zhonglj@cnu.edu.cn.
        \IEEEcompsocthanksitem G.-M. Muntean is with School of Electronic Engineering, Dublin City University, Glasnevin, Dublin 9, Ireland.
        E-mail: gabriel.muntean@dcu.ie.
        \IEEEcompsocthanksitem E. Diao is with DreamSoul, P. R. China. 
        E-mail: diao\_em@hotmail.com
        \IEEEcompsocthanksitem The corresponding author is Tian Du.
    }
}
\fi

\maketitle

\setlength{\textfloatsep}{6pt}
\setlength{\floatsep}{6pt}
\setlength{\intextsep}{6pt}

\setlength{\abovecaptionskip}{3pt}
\setlength{\belowcaptionskip}{0pt}

\begin{abstract}

Federated Learning (FL) faces challenges from client data heterogeneity and resource-constrained mobile devices, which can degrade model accuracy. Personalized Federated Learning (PFL) addresses this issue by adapting shared global knowledge to local data distributions.
A promising approach in PFL is model decoupling, which separates the model into global and personalized parameters, raising the key question of which parameters should be personalized to balance global knowledge sharing and local adaptation.
In this paper, we propose a Federated Optimal Brain Personalization ({\tt FedOBP}) algorithm with a quantile-based thresholding mechanism and introduce an element-wise importance score.
This score extends Optimal Brain Damage (OBD) pruning theory by incorporating a federated approximation of the first-order derivative in the Taylor expansion to evaluate the importance of each parameter for personalization. Moreover, we move the metric computation originally performed on clients to the server side, to alleviate the burden on resource-constrained mobile devices.
To the best of our knowledge, this is the first work to bridge classical saliency-based pruning theory with federated parameter decoupling, providing a rigorous theoretical justification for selecting personalized parameters based on their sensitivity to local loss landscapes.
Extensive experiments demonstrate that {\tt FedOBP} outperforms state-of-the-art methods across diverse datasets and heterogeneity scenarios, while requiring personalization of only a very small number of personalized parameters.
Our code is available \href{https://github.com/uglyghost/FedOBP}{here}.
\end{abstract}


\section{Introduction}
\label{main:introduction}

As deep models scale with growing data, efficient large-scale training has become a key challenge for distributed learning across mobile devices and organizations. Federated Learning (FL)~\citep{mcmahan2017communication} enables clients to collaboratively train a shared global model while keeping their local data private. In practice, limited bandwidth, energy constraints, and heterogeneity make partial participation common, so only a subset of clients join each communication round. Specifically, client data are typically non-independent and non-identically distributed (non-IID), which can induce client drift~\citep{karimireddy2020scaffold,Gao_2022_CVPR}, where local updates deviate from the global objective and degrade convergence stability and accuracy~\citep{karimireddy2020scaffold}. A practical federated scheme should maintain high accuracy under non-IID data and low overhead under partial participation of resource-constrained clients.

Personalized Federated Learning (PFL) enables individual clients to maintain customized models tailored to their local data while sharing knowledge across clients~\citep{11398744,11075614,chen2023metafed,yao2023f,wu2024fedcache,11329154,11194117,li2021ditto,zhang2021personalized,chen2022bridging,10933559,10737242,yang2023personalized,zhang2024cpper,arivazhagan2019federated,deng2020adaptive,liang2020think,collins2021exploiting,ohfedbabu,xu2023personalized,tan2023pfedsim,chen2024towards,mclaughlinpersonalized,yang2023dynamic,10472079,tamirisa2024fedselect}. 
Existing PFL methods primarily involve techniques such as knowledge distillation to transfer teacher-student information~\citep{11398744,11075614,chen2023metafed,yao2023f,wu2024fedcache}, loss regularization to constrain local deviations~\citep{11329154,11194117,li2021ditto,zhang2021personalized,chen2022bridging}, and similarity-based aggregation to cluster kindred clients~\citep{10933559,10737242,yang2023personalized,zhang2024cpper}.
A representative PFL approach is model decoupling~\citep{arivazhagan2019federated,deng2020adaptive,liang2020think,collins2021exploiting,ohfedbabu,xu2023personalized,tan2023pfedsim,chen2024towards,mclaughlinpersonalized}, which divides the neural network model into a globally shared feature extractor and a locally personalized prediction head. 
Parameter decoupling~\citep{yang2023dynamic,10472079,tamirisa2024fedselect} as an extension, provides a finer-grained decomposition by dividing model parameters into element-wise global and personalized subsets. However, maintaining personalization incurs additional computational overhead.
Complex personalization parameter selection increases local memory usage, while excessive personalization reduces the shared backbone, weakening the collaborative benefits. Therefore, a critical challenge of parameter decoupling is identifying which parameters should be personalized. 


Studies like FedSelect~\citep{tamirisa2024fedselect} and PSPFL~\citep{10472079} identify personalized parameters through significant parameter changes (e.g., accumulated gradients) during local training. When client participation is less than 1, these methods can only reuse parameters when the client is selected again, leading to metric delays. FedDPA~\citep{yang2023dynamic} addresses this by selecting personalized parameters based on Fisher information, computed before local training to avoid delays. However, calculating Fisher value introduces considerable local overhead. Moreover, both approaches typically require a large proportion of personalized parameters for optimal performance, which limits the ability of clients to leverage shared global knowledge. An ideal PFL solution should be friendly to partial client participation, have low/no local computational overhead, and achieve high performance with only a small number of personalized parameters.

In this paper, we propose {\tt FedOBP}, a novel PFL framework that emphasizes element-wise importance decoupling based on the Optimal Brain Damage (OBD) theory. To identify personalized parameters, we extend OBD to the federated setting and derive a theoretically grounded and interpretable importance scoring method. By strictly identifying critical weights for local adaptation, our approach achieves superior performance under non-IID data distributions, while requiring very few personalized parameters.
Moreover, the computation of the OBD-based importance metric is performed on the server side, thereby alleviating the burden on resource-constrained mobile devices.
The main contributions of this work are as follows:
\begin{itemize}
    \item We introduce a fine-grained parameter decoupling framework equipped with a dynamic quantile-based thresholding mechanism. This approach isolates a sparse subset of client-specific parameters to maintain locally, enabling precise personalization without compromising the generalization capability of the shared global backbone.
    \item We derive {\tt FedOBP}, a theoretically grounded scoring function for identifying critical personalized parameters. By extending the classic OBD theory~\citep{lecun1989optimal} to the federated setting, {\tt FedOBP} incorporates a federated approximation of first-order Taylor expansion terms, rigorously quantifying the impact of individual parameters on local loss reduction. The computation of the OBD-based metric is performed on the server side to reduce the downlink communication overhead.
    \item Extensive empirical evaluations show that our method {\tt FedOBP} achieves state-of-the-art performance with an extremely low personalization ratio ($<0.5\%$), outperforming gradient-based and Fisher-based scoring methods.
    This confirms that {\tt FedOBP} ensures both high efficiency and effectiveness, delivering superior accuracy with minimal deviation from the global model structure.
\end{itemize}

\section{Related Works}
\label{main:related_work}
\subsection{Personalized Federated Learning} 
PFL has been studied from multiple perspectives, focusing on addressing data heterogeneity and enhancing model performance.
For instance, regularization-based methods like pFedMe~\citep{11194117} and Ditto~\citep{li2021ditto} formulate PFL as a bi-level optimization problem, adding proximal constraints to balance local fitting with global alignment. 
Beyond these architectural strategies, APFL~\citep{deng2020adaptive} allows clients to train local models while contributing to the global model by adaptively combining local and global parameters.
FedPer~\citep{arivazhagan2019federated} introduces a layer-wise decoupling design, separating base and personalized layers to address data heterogeneity. 
LG-FedAvg~\citep{liang2020think} takes the opposite approach by training the feature extractor locally and the classifier globally to mitigate the effects of data heterogeneity. In contrast, 
FedRep~\citep{collins2021exploiting} trains the feature extractor globally while training the classifier locally to tackle heterogeneity issues. 
pFedFDA~\citep{mclaughlinpersonalized} addresses the limitations of transitional layer-wise model decoupling, particularly the bias-variance trade-off in classifier training, which relies solely on local datasets. It also views classifier representation learning as a generative modeling task, training representations based on the global feature distribution. 
FLOCO~\citep{grinwaldfederated} leverages linear mode connectivity to identify a linearly connected low-loss region within the parameter space of neural networks. 
This approach allows clients to personalize their local models within designated subregions, while simultaneously collaborating to train a global model.
FLUTE~\citep{liu2024federated} considers federated representation learning under-parameterized regime, which integrates low-rank matrix approximation techniques with FL analysis. Some other methods~\citep{chen2022bridging,tan2023pfedsim} treat the model as a global feature extractor with a personalized classifier head. Building on this design, FedRoD~\citep{chen2022bridging} maintains both a personalized head (P-Head) for each client and a global head (G-Head), which are trained with separate loss functions.


\subsection{Personalized Parameter Importance Score} 
Building on the parameter decoupling, some approaches use element-wise scoring strategies to identify personalized parameters. For instance, FedSelect~\citep{tamirisa2024fedselect} adopts an iterative subnetwork discovery mechanism inspired by the Lottery Ticket Hypothesis, incrementally selecting parameters that exhibit significant local updates to form personalized subnetworks. Similarly, PSPFL~\citep{10472079} prioritizes parameters based on training dynamics to balance transmission overhead with model accuracy. In contrast, FedDPA~\citep{yang2023dynamic} employs Fisher information-based scoring to estimate the sensitivity of each parameter, specifically retaining parameters with high Fisher values locally to preserve client-specific features while sharing less sensitive weights. 
SeqFedEDT~\citep{du2025seqfededt} extends element-wise decoupling to sequential FL by introducing a quantile-based thresholding mechanism to separate shared and personalized subsets, comprehensively exploring three scoring metrics.

\subsection{Optimal Brain Damage}
Optimal Brain Damage~\citep{lecun1989optimal} and Optimal Brain Surgery (OBS)\citep{hassibi1992second} quantify parameter importance primarily through a second-order Taylor expansion of the loss function. Specifically, these methods approximate the increase in error caused by removing parameters by analyzing the curvature of the loss landscape via the Hessian matrix, operating under the assumption that the network has converged to a local minimum where first-order gradients vanish. OBD has been well-established in theory and validated across various fields, including large language models\citep{ma2023llm,zhang2023loraprune}.
However, its application to PFL remains underexplored due to the complexities of the distributed learning paradigm.
In this work, one of our key contributions is extending OBD by incorporating a federated approximation for personalized parameter selection in PFL.

\section{Method}
\label{method}
\subsection{Preliminary}
\label{method:preliminary}
\textbf{Federated Learning}.
We consider the FL training process to consist of $T$ communication rounds. 
In each round $t$, a subset of clients $\mathcal{C}^t \subseteq \mathcal{C}$ is selected, where $|\mathcal{C}^t| = \gamma \cdot |\mathcal{C}|$, $\gamma\in(0,1]$ denotes the number of participating clients, with $t \in [1, T]$ and the participation rate $\gamma$. 
Initially, the server distributes the initialized model $\bm{\theta}_g^0$ to selected clients. 
These clients then perform local training using their client-specific datasets $\mathcal{D}_i$.
The local objective for each client \( i \in \mathcal{C}^t \) is to minimize the expected loss over its local dataset \( \mathcal{D}_i \):
\begin{equation}
    \underset{\bm{\theta}_i}{\arg \min} \, \mathcal{L} (\bm{\theta}_i;\mathcal{D}_i) = \mathbb{E}_{(x, y) \sim \mathcal{D}_i} [\ell(\bm{\theta}_i; (x, y))],
    \label{obj:local}
\end{equation}
where \( \ell(\bm{\theta}_i; (x, y)) \) is the loss function for a sample $(x,y)$ with parameters $\bm{\theta}_i$, and \( \mathcal{L}(\bm{\theta}_i;\mathcal{D}_i) \) is the expected loss on dataset \( \mathcal{D}_i \).
After local training, clients send their updated models $\{{\bm{\theta}}_i^t\}_{i \in \mathcal{C}^t}$ back to the server for the global aggregation:
\begin{equation}
    \bm{\theta}_g^t = \sum_{i \in \mathcal{C}^t} \frac{m_i^t}{m^t} {\bm{\theta}}_i^t,
    \label{eq:fedavg_aggregation}
\end{equation}
where $m^{t}=\sum_{i\in\mathcal{C}^t}{m_i^{t}}$ and $m_i^{t}$ is the local sample count of client $i$ at $t$. The aggregated parameters $\bm{\theta}_g^t$ are then distributed to the local clients for the next communication round. Therefore, the global objective can be expressed as:
\begin{equation}
    \begin{aligned}
    \underset{\bm{\theta}}{\arg \min} \, \mathcal{L}(\bm{\theta}_g;\mathcal{D}) 
        & = \frac{1}{|\mathcal{C}|} \sum_{i \in \mathcal{C}} \mathbb{E}_{(x, y) \sim \mathcal{D}_i} [\ell(\bm{\theta}_g; (x, y))],\\
        & = \mathbb{E}_{(x, y) \sim \mathcal{D}} [\ell(\bm{\theta}_g; (x, y))],
    \end{aligned}
    \label{obj:global}
\end{equation}
where $\mathcal{L}(\bm{\theta}_g;\mathcal{D})$ represents the expected loss over the entire dataset $\mathcal{D} = \{\mathcal{D}_i\}_{i\in \mathcal{C}}$ across all clients.

\textbf{Model Decoupling}.
Model Decoupling addresses data distribution heterogeneity by selecting a client-specific personalized subset \(\bm{u}^t_i\) from the previous local model $\bm{\theta}_i^{t-1}$ and choosing a globally shared subset \(\bm{v}^t_i\) from the global model parameters \(\bm{\theta}^t_g\). Client $i$ combines the client-specific personalized subset \(\bm{u}^t_i\) with the globally shared subset \(\bm{v}^t_i\) to create the merged model $\tilde{\bm{\theta}}_i^t=\{\bm{u}^t_i,\bm{v}^t_i\}$.
Let $\mathcal{K}$ denote the set of all parameter indices, such that $\bm{\theta}^t_g = \{\theta^{t,k}_g\}_{k \in \mathcal{K}}$. For subsets $\bm{u}^t_i$ and $\bm{v}^t_i$, their corresponding parameter index sets are $\mathcal{K}(\bm{u}^t_i)$ and $\mathcal{K}({\bm{v}}^t_i)$ respectively, where $\bm{u}^t_i = \{\theta_i^{t-1,k}\}_{k\in\mathcal{K}({\bm{u}}^t_i)}$ and $\bm{v}^t_i = \{\theta_g^{t,k}\}_{ k\in\mathcal{K}({\bm{v}}^t_i)}$. It is important to note that the element-wise parameter decoupling for each parameter $k$ can vary across clients $i$ and communication rounds $t$.
While \(\bm{u}^t_i\) is updated exclusively using the client’s local dataset \(\mathcal{D}_i\), \(\bm{v}^t_i\) is involved in both local updates and global parameter aggregation.
The local objective function for PFL with parameter decoupling can be expressed as:
\begin{equation}
    \underset{\{\tilde{\bm{\theta}}_i^t\}_{i\in\mathcal{C}}}{\arg \min} \left\{ \frac{1}{|\mathcal{C}|} \sum_{i \in \mathcal{C}} \mathcal{L}(\tilde{\bm{\theta}}_i^t; \mathcal{D}_i) \right\},
    \label{PFL_object}
\end{equation}
where \( \mathcal{L}(\tilde{\bm{\theta}}_i^t;\mathcal{D}_i) = \mathbb{E}_{(x, y) \sim \mathcal{D}_i} [\ell(\tilde{\bm{\theta}}_i^t; (x, y))] \) represents the expected loss for client \( i \) with its decoupled parameters. \( \ell(\tilde{\bm{\theta}}_i^t; (x, y)) \) is the loss function for a sample $(x, y)$ computed using the client-specific model \( \tilde{\bm{\theta}}_i^t \).

\textbf{Parameter Importance Score}.
PFL with parameter decoupling methods~\citep{yang2023dynamic,10472079,tamirisa2024fedselect} identify the personalized parameter set $\bm{u}_i^{t}$ and the globally shared parameter set $\bm{v}_i^{t}$ based on an element-wise parameter importance score $I(\cdot)$.

FedSelect~\citep{tamirisa2024fedselect} and PSPFL~\citep{10472079} propose using local updates from the pre-trained merged model $\tilde{\bm{\theta}}_i^{t-1}$ in the previous round to compute the importance score of each parameter $\theta_i^{t-1,k},k\in\mathcal{K}$. This approach relies on multiple local updates and gradient computations during local training. Specifically, the gradient-based importance score $I_G(\cdot)$ is calculated as the absolute difference between the merged model \(\tilde{\theta}_i^{t-1,k}\) and the locally updated \(\theta_i^{t-1,k}\) as follows:
\begin{align}
    & I_{G}\left(\theta_i^{t-1,k}; \mathcal{D}_i\right) 
    = \left|{\tilde{\theta}}_i^{t-1,k} - \theta_i^{t-1,k} \right|.
    \label{Gradient}
\end{align}
However, gradient-based importance scores can only be determined after local training in the previous communication round $t-1$. This limitation means that the score can only be used if the client is selected again in subsequent rounds, introducing a delay issue when the participation rate $\gamma < 1$.

FedDPA~\citep{yang2023dynamic} takes a different approach by utilizing Fisher information to determine importance for personalization. The Fisher information-based importance score $I_F(\cdot)$ is defined as:
\begin{align}
    I_{F}\left({\theta}_i^{t-1,k}; \mathcal{D}_i\right) = \left( \frac{\partial \mathcal{L}({\theta}_i^{t-1,k}, \mathcal{D}_i)}{\partial {\theta}_i^{t-1,k}} \right)^2.
    \label{Fisher}
\end{align}
Unlike gradient-based methods, the Fisher information-based approach can be applied before local training and does not suffer from the delay issue. However, it still requires gradient computation in order to calculate the Fisher information for each parameter $\{\theta^{t-1,k}\}_{k\in\mathcal{K}}$ with the local dataset $\mathcal{D}_i$, which introduces additional local computational overhead.
Furthermore, we observe that both the gradient-based and Fisher information-based methods require a relatively large proportion of personalized parameters to achieve optimal performance. This requirement for a large proportion of personalized parameters limits the ability of local models to effectively leverage shared knowledge across clients, thereby reducing the overall benefits of collaboration.


\begin{figure*}[t]
    \centering
    \includegraphics[width=0.8\linewidth]{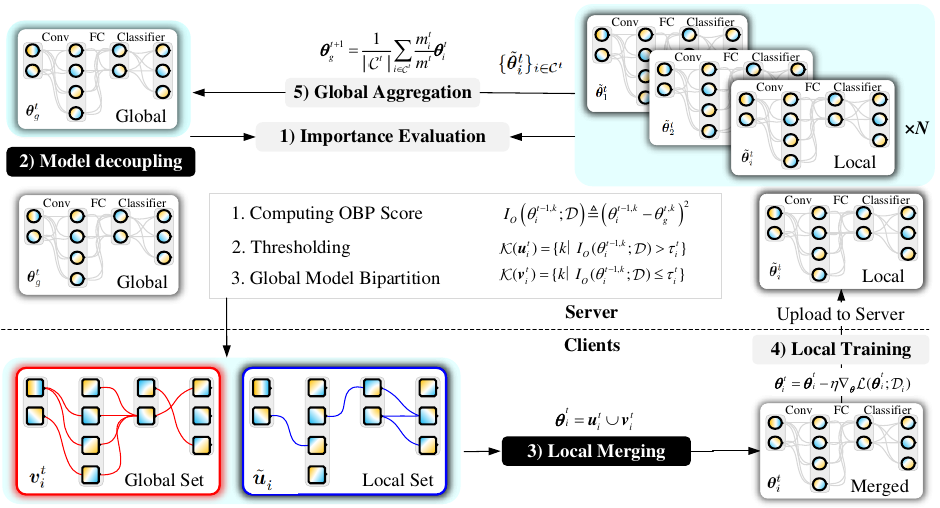}
    \caption{Overview of \texttt{FedOBP}. 
    The server computes the Federated OBP parameter importance $I_{O}(\bm{\theta}_i^{t-1};\mathcal{D})$ for each selected client based on the uploaded local model and the aggregated global model $\bm{\theta}_g^t$, and determines the personalized parameter subset $\bm{u}_i^t$ and the globally shared parameter subset $\bm{v}_i^t$ using a quantile-based thresholding mechanism. The server then sends $\bm{v}_i^t$ to client $i$. Client $i$ merges the globally shared parameters $\bm{v}_i^t$ with the personalized parameters $\bm{u}_i^t$ to form the merged model $\tilde{\bm{\theta}}_i^{t}$. It then performs local training on $\tilde{\bm{\theta}}_i^{t}$ and uploads the updated model $\bm{\theta}_i^t$ to the server for global aggregation.}
    \label{fig:FedOBP}
    \vspace{-1.5em}
\end{figure*}

\subsection{\texttt{FedOBP} Algorithm}
\label{method:FedOBP}

The workflow of the {\tt FedOBP} algorithm is illustrated in Figure~\ref{fig:FedOBP}.
\texttt{FedOBP} follows the general framework of standard PFL, incorporating the Federated OBP parameter importance score function $I_{O}(\cdot)$. To accommodate the limited computational resources of edge devices, the importance metric is computed on the server side rather than on the clients. The pseudocode of the main steps is provided in Algorithm~\ref{alg:FedOBP}.

\textit{1) Importance Evaluation:} 
Based on the previously uploaded local models and the current global model, the server computes the Federated OBP parameter importance
\[
I_O(\bm{\theta}_i^{t-1};\mathcal{D}) = \{I_O(\theta_i^{t-1,k}; \mathcal{D})\}_{k\in\mathcal{K}}
\]
for each selected client $i$. We will provide details of the Federated OBP parameter importance function in Section~\ref{method:OBPimportance}.

\textit{2) Model Decoupling:} 
For each selected client $i$, the server adopts an adaptive thresholding mechanism based on the Federated OBP parameter importance to decouple the personalized set $\bm{u}_i^t$ from the previous local model $\bm{\theta}_i^{t-1}$ and the globally shared subset $\bm{v}_i^t$ from the global model $\bm{\theta}_g^t$. The importance-based partitioning uses a quantile-based thresholding function
\[
f_q: I_O(\bm{\theta}_i^{t-1};\mathcal{D}) \rightarrow \mathbb{R},
\]
which determines a threshold $\tau_i^t$ as follows:
\begin{align}
    \tau_i^t &= f_{q}(I_O(\bm{\theta}_i^{t-1};\mathcal{D})) \notag\\
    &\triangleq \inf\{I_O(\bm{\theta}_i^{t-1};\mathcal{D}) \in \mathbb{R}: F(I_O(\bm{\theta}_i^{t-1};\mathcal{D})) \geq q\}, \notag
\end{align}
where $F(x)$ is the cumulative distribution function of $I_O(\bm{\theta}_i^{t-1};\mathcal{D})$ and $q \in [0,1]$ is the quantile level.

Using this threshold \(\tau_i^t\), the server determines the personalized parameter set $\bm{u}_i^{t} = \{\theta_i^{t-1,k}\}_{k\in\mathcal{K}({\bm{u}}^t_i)}$ and the globally shared parameter set $\bm{v}_i^{t} = \{\theta_g^{t,k}\}_{ k\in\mathcal{K}({\bm{v}}^t_i)}$ with the corresponding index sets as follows:
\begin{align}
    \mathcal{K}({\bm{u}}^t_i) &= \{k \mid I_{O}({\theta}_i^{t-1,k};\mathcal{D}) > \tau_i^t \},\label{MD1} \\
    \mathcal{K}({\bm{v}}^t_i) &= \{k \mid I_{O}({\theta}_i^{t-1,k};\mathcal{D}) \leq \tau_i^t \},
    \label{MD2}
\end{align}
where ${\theta}_i^{t-1,k}$ represents the previous local model parameter at position $k$, and \(\tau_i^t\) is a threshold. The server then transmits the selected global parameter subset $\bm{v}_i^t$ to client $i$. In this way, the computation of the OBD-based importance metric is performed on the server side, alleviating the burden on resource-constrained clients.

Through extensive numerical experiments in Section~\ref{Importance_Scores}, we found that {\tt FedOBP} can achieve strong performance by selecting very few personalized parameters $\bm{u}_i$. Notably, most selected personalized parameters are concentrated in the classifier layer, aligning with theoretical insights from Centered Kernel Alignment (CKA)~\citep{hinton2015distilling}.

\textit{3) Local Merging:} 
Each client \(i\) receives the global parameter subset $\bm{v}_i^{t}$ from the server and obtains the merged model $\tilde{\bm{\theta}}_i^{t}$ by combining $\bm{v}_i^{t}$ with the client-specific personalized parameter subset $\bm{u}_i^{t}$ retained from the previous local model:
\begin{align}
    \tilde{\bm{\theta}}_i^{t} = \bm{u}_i^{t} \cup \bm{v}_i^{t}. \label{merge}
\end{align}

\textit{4) Local Training:} 
Each client $i$ performs local training on the merged model $\tilde{\bm{\theta}}_i^{t}$ with the local dataset $\mathcal{D}_i$ to obtain the locally trained model $\bm{\theta}_i^{t}$, following the update rule:
\begin{equation}
    \bm{\theta}_i^{t} = \tilde{\bm{\theta}}_i^{t} - \eta \nabla_{\bm{\theta}} \mathcal{L}(\tilde{\bm{\theta}}_i^{t};\mathcal{D}_i),
    \label{localTraining}
\end{equation}
where $\eta$ is the learning rate. In FL, such model updates can be performed multiple times. Then, each client $i$ uploads its locally trained model $\bm{\theta}_i^{t}$ to the server.

\textit{5) Global Aggregation:} 
The server aggregates the locally trained models $\{\bm{\theta}_i^{t}\}_{i\in\mathcal{C}^t}$ to compute the globally aggregated model $\bm{\theta}_g^{t+1}$ for the next communication round:
\begin{equation}
    \bm{\theta}_g^{t+1} = \frac{1}{|\mathcal{C}^t|} \sum_{i \in \mathcal{C}^t} \frac{m_i^t}{m^{t}} \bm{\theta}_i^{t}.
    \label{aggregation}
\end{equation}

\subsection{Federated OBP Parameter Importance}
\label{method:OBPimportance}
OBD~\citep{lecun1989optimal,hassibi1992second,molchanov2019importance,zhang2023loraprune,ma2023llm} is a model pruning technique that quantifies the element-wise importance of each parameter \( \theta^{k} \) in a model $\bm{\theta} = \{\theta^k\}_{k\in\mathcal{K}}$ with respect to the loss function $\mathcal{L}(\cdot)$ on dataset $\mathcal{D}$ as:
\begin{align}
    I_{O}(\theta^{k};\mathcal{D}) = |\Delta \mathcal{L}(\theta^{k};\mathcal{D})| = |\mathcal{L}(\theta^{k}_{=0};\mathcal{D}) - \mathcal{L}(\theta^{k};\mathcal{D})|,
    \label{Important}
\end{align}
where \( \mathcal{L}(\theta^{k}_{=0};\mathcal{D}) \) represents the expected loss with parameter $\theta^{k}$ is set to $0$.
The importance $I_{O}(\theta^{k};\mathcal{D})$ of parameter $\theta^k$ at position $k$ can be further expanded using a Taylor series approximation of $\mathcal{L}(\theta^{k}_{=0};\mathcal{D})$ at $\theta^k$ to obtain:
\begin{align}
    & I_{O}(\theta^{k};\mathcal{D}) \notag \\
    & = \left| \frac{\partial \mathcal{L}(\theta^k;\mathcal{D})}{\partial \theta^k} \delta \theta^k + \frac{1}{2} \delta \theta^k H_{kk} \delta \theta^k + \mathcal{O}(\|\theta^k\|^3) \right|,
    \label{Taylor}
\end{align}
where $\delta \theta^k = \theta^{k}_{=0} - \theta^k = - \theta^k$ and \( H_{kk} \) is the diagonal entry of the Hessian matrix, capturing the second-order curvature of \( \mathcal{L}(\cdot) \) with respect to \( \theta^k \).
\( \mathcal{O}(\|\theta^k\|^3) \) denotes higher-order terms.
Classical OBD and OBS methods primarily focus on pruning models after convergence relying on the second order term of the Taylor series approximation while assuming the first order term to be negligible~\citep{lecun1989optimal,hassibi1992second}.

\begin{algorithm}[!]
    \caption{\textbf{\texttt{FedOBP}} ($T$, $\gamma$)}
    \label{alg:FedOBP}
    \SetAlgoLined
    \SetKwInput{KwInput}{Input}
    \SetKwInput{KwOutput}{Output}
    
    \KwInput{Total rounds $T$, participation rate $\gamma$.}
    \KwOutput{Global model $\bm{\theta}_g^T$, local models $\{\bm{\theta}_i^T\}_{i \in \mathcal{C}}$}
    
    Initialize global model $\bm{\theta}_g^0$\ and local models $\bm{\theta}_i^0 \gets \bm{\theta}_g^0$, $\forall i \in \mathcal{C}$\;
    
    \For{$t = 1$ \KwTo $T$}{
        Select clients $\mathcal{C}^t$, where $|\mathcal{C}^t| = \gamma |\mathcal{C}|$\;
        
        \ForEach{$i \in \mathcal{C}^t$}{
            Compute \texttt{FedOBP} parameter importance 
            $\{I_{O}({\theta}_i^{t-1,k};\mathcal{D})\}_{k\in\mathcal{K}}$ based on Eq.~\ref{FedOBP2}\;
            
            Decouple $\{\bm{u}_i^{t}, \bm{v}_i^{t}\}$ based on Eq.~\ref{MD1} and Eq.~\ref{MD2}\;
            
            Send $\bm{v}_i^{t}$ to client $i$\;
        }
        
        \ForEach{$i \in \mathcal{C}^t$ {in parallel}}{
            Retain $\bm{u}_i^t$ from the previous local model $\bm{\theta}_i^{t-1}$\;
            
            Merge model $\tilde{\bm{\theta}}_i^{t} = \bm{u}_i^{t} \cup \bm{v}_i^{t}$\;
            
            Train $\bm{\theta}_i^{t} \gets \tilde{\bm{\theta}}_i^{t} - \eta \nabla_{\bm{\theta}} \mathcal{L}(\tilde{\bm{\theta}}_i^{t}; \mathcal{D}_i)$\;
            
            Upload $\bm{\theta}_i^{t}$ to the server\;
        }
        
        Aggregate $\{\bm{\theta}_i^{t}\}_{i\in\mathcal{C}^t}$ based on Eq.~\ref{aggregation} to obtain $\bm{\theta}_g^{t}$\;
    }
    
    \Return $\bm{\theta}_g^T$, $\{\bm{\theta}_i^T\}_{i \in \mathcal{C}}$\;
\end{algorithm}

\textbf{Federated Optimal Brain Personalization (\texttt{FedOBP})}.
As discussed in Section~\ref{method:preliminary}, global aggregation optimizes the global model $\bm{\theta}_g^t$ by minimizing the global loss~\eqref{obj:global} on the global dataset $\mathcal{D}$. However, due to data heterogeneity across clients, this aggregation can degrade the performance of each personalized model $\bm{\theta}_i^t$ on its corresponding local dataset $\mathcal{D}_i$. 
To maximize the benefits of shared global knowledge, we aim to share most parameters across clients and personalize very few parameters. We choose to identify the local parameters $\theta_i^{t-1,k}$ that are most critical for the global dataset $\mathcal{D}$.

Building on the OBD pruning theory, we introduce the Federated Optimal Brain Personalization score function to assess the importance of each local model parameter ${\theta}_i^{t-1,k}$ with respect to the corresponding global model parameter $\theta_g^{t,k}$ for the global dataset $\mathcal{D}$. We use a Taylor series approximation of $\mathcal{L}(\theta_g^{t,k};\mathcal{D})$ at $\theta_i^{t-1,k}$ to obtain:
\begin{align}
    & I_{O}\left({\theta}_i^{t-1,k};\mathcal{D}\right) 
     = \left|\mathcal{L}\left({\theta}^{t,k}_{g};\mathcal{D}\right) -\mathcal{L}\left({\theta}_i^{t-1,k};\mathcal{D}\right)\right| \\
    &= \Bigg| \frac{\partial \mathcal{L}(\theta_i^{t-1,k};\mathcal{D})}{\partial \theta_i^{t-1,k}} \delta \theta_i^{t-1,k} + \frac{1}{2} \delta \theta_i^{t-1,k} H_{kk} \delta \theta_i^{t-1,k} \notag\\
    & \quad + \mathcal{O}(\|\theta^{t-1,k}\|^3) \Bigg|,
\end{align}
where $\delta \theta_i^{t-1,k} = \theta^{t,k}_{g} - \theta_i^{t-1,k}$. In classical OBD, the importance of parameters is measured by the loss difference between pruned and unpruned parameters. In the proposed \texttt{FedOBP}, the importance of global parameters with respect to the global dataset is measured by the loss difference between local and global parameters.

\texttt{FedOBP} applies parameter importance analysis during the FL training phase, where the first order term often dominates the second order term in magnitude. \texttt{FedOBP} approximates parameter importance using only the first order term~\citep{molchanov2019importance,zhang2023loraprune}:
\begin{align}
    I_{O}\left({\theta}_i^{t-1,k};\mathcal{D}\right) 
    & \approx \left| \frac{\partial \mathcal{L}(\theta_i^{t-1,k};\mathcal{D})}{\partial \theta_i^{t-1,k}} \cdot (\theta^{t,k}_{g} - {\theta}_i^{t-1,k})\right| \label{FedOBP1}
\end{align}
Meanwhile, we interpret the global aggregation process in Eq.~\ref{aggregation} as performing a single update step for the local models $\{{\bm{\theta}}^{t-1}_i\}_{i\in\mathcal{C}^{t}}$ on the global dataset $\mathcal{D}$.
Consequently, the gradient descent formulation of the global aggregation for each parameter $\theta_i^{t-1,k}$ can be expressed as:
\begin{align}
    \theta_g^{t,k} \approx \theta_i^{t-1,k} &- \eta \frac{\partial \mathcal{L}(\theta_i^{t-1,k};\mathcal{D})}{\partial \theta_i^{t-1,k}}.
\end{align}
According to the classical FedAvg, multi-step cumulative gradient updates provide more accurate parameter improvements than a single-step gradient descent update~\citep{mcmahan2017communication}.
A well-known federated optimization approach, Adaptive Federated Optimization (AFO)~\citep{reddiadaptive}, also interprets global aggregation as one step of gradient descent. Therefore, the federated gradient can be regarded as global-local parameter update as follows:
\begin{align}
    \frac{\partial \mathcal{L}(\theta_i^{t-1,k};\mathcal{D})}{\partial \theta_i^{t-1,k}} &\approx \theta_i^{t-1,k} - \theta_g^{t,k}.
    \label{FSGD}
\end{align}
Unlike AFO, which treats sequential global aggregations $({\theta}_g^{t-1,k} - \theta_g^{t,k})$ as ``gradient'', we regard the federated global-local parameter update $({\theta}_i^{t-1,k} - {\theta}_g^{t,k})$ as an approximation to the gradient. Therefore, we can further approximate Eq.~\ref{FedOBP1} by replacing the gradient term with the parameter update $({\theta}_i^{t-1,k} - \theta_g^{t,k})$ in Eq.~\ref{FSGD} and obtain the \texttt{FedOBP} score:
\begin{align}
    I_{O}\left({\theta}_i^{t-1,k};\mathcal{D}\right) 
    \triangleq \left({\theta}_i^{t-1,k} - \theta_g^{t,k} \right)^2.
    \label{FedOBP2}
\end{align}

\section{Interpreting FedAvg as a Single Global Gradient Step}

In this section, we demonstrate that the {\tt FedAvg} aggregation at round $t$ can be approximated as a single gradient descent step on the global objective function $\mathcal{L}(\theta;\mathcal{D})$, evaluated at the local model parameter $\theta_i^{t-1}$. 
This derivation provides the theoretical justification for approximating the global gradient using the difference between local and global parameters.

\subsection{Setup and Definitions}

Let $\theta_g^{t-1}$ be the global model distributed at the start of round $t$.
Each client $i$ initializes $\theta_{i,0}^{t-1} = \theta_g^{t-1}$ and performs $E$ steps of local SGD on its local dataset $\mathcal{D}_i$ on its local loss $\mathcal{L}(\theta; \mathcal{D}_i)$. The update rule for the $s$-th step ($s \in \{1, \dots, E\}$) is:
\begin{equation}
    \theta_{i,s}^{t-1}
    =
    \theta_{i,s-1}^{t-1}
    -
    \eta \nabla_{\theta}\mathcal{L}(\theta_{i,s-1}^{t-1};\mathcal{D}_i), 
    \quad s=1,\dots,E.
    \label{eq:local_sgd_corrected}
\end{equation}
The final local model for client $i$ is denoted as $\theta_i^{t-1} \triangleq \theta_{i,E}^{t-1}$.
The server then aggregates these models to produce the global model for the next round:
\begin{equation}
    \theta_g^t = \sum_{j \in \mathcal{C}^t} \frac{m_j^t}{m^t} \theta_j^{t-1}.
    \label{eq:fedavg_corrected}
\end{equation}
Our goal is to analyze the deviation $\theta_g^t - \theta_i^{t-1}$ to approximate the global gradient as follows:
\begin{equation}
\theta_g^t - \theta_i^{t-1}
=
\sum_{j \in \mathcal{C}^t}
\frac{m_j^t}{m^t}
(\theta_j^{t-1} - \theta_i^{t-1}).
\label{eq:global_local_diff_corrected}
\end{equation}

\subsection{Unrolling Local SGD}
We begin by decomposing the deviation between the aggregated global model $\theta_g^{t-1}$ and the local model $\theta_{i,s}^{t-1}$ of a specific client $i$ after $s$ SGD steps:
\begin{equation}
\theta_{j,s}^{t-1}
=
\theta_g^{t-1}
-
\eta \sum_{\tau=1}^{s}
\nabla_{\theta}\mathcal{L}(\theta_{j,\tau-1}^{t-1};\mathcal{D}_j).
\label{eq:unroll_j_corrected}
\end{equation}
Similarly, for client $i$:
\begin{equation}
\theta_{i,s}^{t-1}
=
\theta_g^{t-1}
-
\eta \sum_{\tau=1}^{s}
\triangledown_{\theta}\mathcal{L}(\theta_{i,\tau-1}^{t-1};\mathcal{D}_i).
\label{eq:unroll_i_corrected}
\end{equation}
Subtracting Eq. \eqref{eq:unroll_i_corrected} from Eq. \eqref{eq:unroll_j_corrected}:
\begin{align}
& \theta_{j,s}^{t-1} - \theta_{i,s}^{t-1}  \notag \\
& =
-\eta
\sum_{\tau=1}^{s}
\Big(
\nabla_{\theta}\mathcal{L}(\theta_{j,\tau-1}^{t-1};\mathcal{D}_j)
-
\nabla_{\theta}\mathcal{L}(\theta_{i,\tau-1}^{t-1};\mathcal{D}_i)
\Big).
\end{align}
Setting $s=E$ gives the difference between the final local models of client $i$ and client $j$:

\begin{equation}
\boxed{
\begin{aligned}
& \theta_j^{t-1} - \theta_i^{t-1} \\
& =
-\eta
\sum_{\tau=1}^{E}
\Big(
\nabla_{\theta}\mathcal{L}(\theta_{j,\tau-1}^{t-1};\mathcal{D}_j)
-
\nabla_{\theta}\mathcal{L}(\theta_{i,\tau-1}^{t-1};\mathcal{D}_i)
\Big)
\end{aligned}
}
\label{eq:B1_final_corrected}
\end{equation}

\subsection{Parameter Drift Bounds}
We bound how far the local models on different clients can drift during the $E$ steps of local SGD in each round to facilitate the approximation in the subsequent steps.

\begin{assumption}[Bounded Gradients]
For all clients $i$ and all parameters $\theta$,
\[
\|\nabla_{\theta}\mathcal{L}(\theta;\mathcal{D}_i)\| \le G.
\]
\end{assumption}

\subsection*{C.1. Drift between clients $i$ and $j$ at step $s$}

At the beginning of round $t$, all selected clients start from the same global model, i.e., $\theta_{i,0}^{t-1} = \theta_{j,0}^{t-1} = \theta_g^{t-1}$.
The updates for clients $i$ and $j$ at step $s$ are:
\[
\theta_{i,s}^{t-1}
=
\theta_{i,s-1}^{t-1}
-
\eta\,\nabla_{\theta}\mathcal{L}(\theta_{i,s-1}^{t-1};\mathcal{D}_i),
\]
\[
\theta_{j,s}^{t-1}
=
\theta_{j,s-1}^{t-1}
-
\eta\,\nabla_{\theta}\mathcal{L}(\theta_{j,s-1}^{t-1};\mathcal{D}_j).
\]

Subtracting these updates yields:
\begin{align}
\theta_{j,s}^{t-1} - \theta_{i,s}^{t-1} & =
(\theta_{j,s-1}^{t-1} - \theta_{i,s-1}^{t-1}) \notag
\\
& 
-
\eta\Big(
\nabla\mathcal{L}(\theta_{j,s-1}^{t-1};\mathcal{D}_j)
-
\nabla\mathcal{L}(\theta_{i,s-1}^{t-1};\mathcal{D}_i)
\Big). \notag
\end{align}

Taking norms and applying the bounded gradient assumption ($ \|\nabla\mathcal{L}\| \le G$):
\begin{align}
&\Big\|\nabla\mathcal{L}(\theta_{j,s-1}^{t-1};\mathcal{D}_j)
-
\nabla\mathcal{L}(\theta_{i,s-1}^{t-1};\mathcal{D}_i)\Big\|
\notag \\
&\le
\|\nabla\mathcal{L}(\cdot;\mathcal{D}_j)\|
+
\|\nabla\mathcal{L}(\cdot;\mathcal{D}_i)\|
\le 2G.
\notag
\end{align}

Hence, the recurrence relation for the distance is:
\[
\|\theta_{j,s}^{t-1} - \theta_{i,s}^{t-1}\|
\le
\|\theta_{j,s-1}^{t-1} - \theta_{i,s-1}^{t-1}\|
+
2\eta G.
\]

Since the initial difference is zero ($\|\theta_{j,0}^{t-1} - \theta_{i,0}^{t-1}\| = 0$), unrolling for $s$ steps yields:
\begin{equation}
\boxed{
\|\theta_{j,s}^{t-1} - \theta_{i,s}^{t-1}\|
\le
2 \eta G s.
}
\tag{C.1}
\end{equation}

\subsection*{C.2. Drift of client $i$ from intermediate iterate to final iterate}

We next bound the distance between an intermediate parameter $\theta_{i,s}^{t-1}$ and the final local parameter $\theta_i^{t-1} = \theta_{i,E}^{t-1}$.
Using the SGD update rule, the difference is the sum of subsequent gradients:
\[
\theta_i^{t-1} - \theta_{i,s}^{t-1}
=
-\eta \sum_{\tau=s+1}^{E}
\nabla_{\theta}\mathcal{L}(\theta_{i,\tau-1}^{t-1};\mathcal{D}_i).
\]

Taking norms and applying bounded gradients:
\[
\begin{aligned}
\|\theta_{i,s}^{t-1} - \theta_i^{t-1}\|
&\le
\eta \sum_{\tau=s+1}^E \|\nabla\mathcal{L}(\cdot;\mathcal{D}_i)\| \\
&\le
\eta G (E-s) \\
&\le
\eta G E.
\end{aligned}
\]

Thus,
\begin{equation}
\boxed{
\|\theta_{i,s}^{t-1} - \theta_i^{t-1}\|
\le
\eta G E.
}
\tag{C.2}
\end{equation}

\subsection*{C.3. Combined drift bound}

Using the triangle inequality and combining (\textit{C.1}) and (\textit{C.2}):
\[
\|\theta_{j,s}^{t-1} - \theta_i^{t-1}\|
\le
\|\theta_{j,s}^{t-1} - \theta_{i,s}^{t-1}\|
+
\|\theta_{i,s}^{t-1} - \theta_i^{t-1}\|.
\]

Substituting (\textit{C.1}) and (\textit{C.2}) and using $s \le E$:
\begin{equation}
\boxed{
\|\theta_{j,s}^{t-1} - \theta_i^{t-1}\|
\le
3 \eta G E.
}
\label{eq:drift_combined_corrected}
\end{equation}

\subsection{Smoothness-Based Gradient Alignment}

\begin{assumption}[Smoothness]
We assume the loss function $\mathcal{L}(\theta;\mathcal{D}_j)$ is $L$-smooth for all $j$:
\[
\|\nabla_{\theta}\mathcal{L}(x;\mathcal{D}_j)
-
\nabla_{\theta}\mathcal{L}(y;\mathcal{D}_j)\|
\le
L\|x-y\|.
\]
\end{assumption}

Using the combined drift bound Eq. \eqref{eq:drift_combined_corrected}, we can approximate the gradient at the intermediate step $s$ using the gradient at the final local model $\theta_i^{t-1}$:
\begin{align}
\nabla_{\theta}\mathcal{L}(\theta_{j,s}^{t-1};\mathcal{D}_j)
&=
\nabla_{\theta}\mathcal{L}(\theta_i^{t-1};\mathcal{D}_j)
+
\delta_{j,s},
\\
\|\delta_{j,s}\|
&\le 3 L G \eta E.
\end{align}

Similarly for client $i$:
\begin{align}
\nabla_{\theta}\mathcal{L}(\theta_{i,s}^{t-1};\mathcal{D}_i)
&=
\nabla_{\theta}\mathcal{L}(\theta_i^{t-1};\mathcal{D}_i)
+
\delta_{i,s},
\\
\|\delta_{i,s}\|
&\le 3 L G \eta E.
\end{align}

Thus, the difference in gradients in Eq. \eqref{eq:B1_final_corrected} becomes:
\begin{align}
& \nabla_{\theta}\mathcal{L}(\theta_{j,s-1}^{t-1};\mathcal{D}_j)
-
\nabla_{\theta}\mathcal{L}(\theta_{i,s-1}^{t-1};\mathcal{D}_i) \notag \\
& =
\big(
\nabla_{\theta}\mathcal{L}(\theta_i^{t-1};\mathcal{D}_j)
-
\nabla_{\theta}\mathcal{L}(\theta_i^{t-1};\mathcal{D}_i)
\big)
+
(\delta_{j,s} - \delta_{i,s}).
\end{align}

Summing over $s=1 \dots E$ gives:

\begin{equation}
    \boxed{
\begin{aligned}
&\sum_{s=1}^{E}
\Big[
\nabla_{\theta}\mathcal{L}(\theta_{j,s-1}^{t-1};\mathcal{D}_j)
-
\nabla_{\theta}\mathcal{L}(\theta_{i,s-1}^{t-1};\mathcal{D}_i)
\Big]
\\
&= E\Big[
\nabla_{\theta}\mathcal{L}(\theta_i^{t-1};\mathcal{D}_j)
-
\nabla_{\theta}\mathcal{L}(\theta_i^{t-1};\mathcal{D}_i)
\Big]
+ O(\eta E^2).
\end{aligned}
}
\label{eq:aligned_sum_corrected}
\end{equation}

\subsection*{E. Substitute Into Aggregation}

We now substitute Eq. \eqref{eq:aligned_sum_corrected} back into the expression for $\theta_g^t - \theta_i^{t-1}$. 
First, substituting into Eq. \eqref{eq:B1_final_corrected} yields:
\begin{align}
\theta_j^{t-1} - \theta_i^{t-1} = -\eta E 
& \Big( \nabla_{\theta}\mathcal{L}(\theta_i^{t-1};\mathcal{D}_j) \notag \\
& - \nabla_{\theta}\mathcal{L}(\theta_i^{t-1};\mathcal{D}_i) \Big) 
 + O(\eta^2 E^2).
\end{align}

Next, substitute this into Eq. \eqref{eq:global_local_diff_corrected}:
\begin{align}
\theta_g^t - \theta_i^{t-1} \notag
= 
\sum_{j \in \mathcal{C}^t} \frac{m_j^t}{m^t} 
& \Big[ -\eta E \Big( \nabla_{\theta}\mathcal{L}(\theta_i^{t-1};\mathcal{D}_j)  \\
& - \nabla_{\theta}\mathcal{L}(\theta_i^{t-1};\mathcal{D}_i) \Big) + O(\eta^2 E^2) \Big] \notag \\
=
-\eta E
\sum_{j \in \mathcal{C}^t}
\frac{m_j^t}{m^t}
& \Big(
\nabla_{\theta}\mathcal{L}(\theta_i^{t-1};\mathcal{D}_j) \notag \\
& - \nabla_{\theta}\mathcal{L}(\theta_i^{t-1};\mathcal{D}_i) \Big) + O(\eta^2 E^2).
\label{eq:E1_corrected}
\end{align}

Using the definition of the global loss gradient $\nabla_{\theta}\mathcal{L}(\theta;\mathcal{D}) = \sum_{j} \frac{m_j^t}{m^t} \nabla_{\theta}\mathcal{L}(\theta;\mathcal{D}_j)$, we have:
\[
\sum_{j \in \mathcal{C}^t} \frac{m_j^t}{m^t}
\nabla_{\theta}\mathcal{L}(\theta_i^{t-1}; \mathcal{D}_j)
=
\nabla_{\theta}\mathcal{L}(\theta_i^{t-1};\mathcal{D}).
\]
Note that the term $\nabla_{\theta}\mathcal{L}(\theta_i^{t-1};\mathcal{D}_i)$ does not depend on $j$, so its weighted sum remains the same. Therefore:

\begin{equation}
    \boxed{
\begin{aligned}
\theta_g^t - \theta_i^{t-1}
& =
-\eta E
\Big[
\nabla_{\theta}\mathcal{L}(\theta_i^{t-1};\mathcal{D})
-
\nabla_{\theta}\mathcal{L}(\theta_i^{t-1};\mathcal{D}_i)
\Big] \\
& +
O(\eta^2 E^2).
\end{aligned}
}
\label{eq:E_final_corrected}
\end{equation}

\subsection*{F. Local Approximate Stationarity on $\mathcal{D}_i$}

The term 
$\nabla_{\theta}\mathcal{L}(\theta_i^{t-1};\mathcal{D}_i)$
in Eq. \eqref{eq:E_final_corrected} is the local gradient evaluated at the
parameter $\theta_i^{t-1}$, which is already the result of $E$ local
SGD updates during round $t$.  
Since $\theta_i^{t-1}$ has been explicitly trained to reduce the local
loss $\mathcal{L}(\cdot;\mathcal{D}_i)$, its local gradient is expected
to be small.  We formalize this as follows:

\begin{assumption}[Local Approximate Stationarity]
Each client outputs a locally updated model $\theta_i^{t-1}$ satisfying
\[
\big\|\nabla_{\theta}\mathcal{L}(\theta_i^{t-1};\mathcal{D}_i)\big\|
\le \varepsilon_i,
\]
where $\varepsilon_i$ is the residual gradient remaining after the
client's local optimization.  We assume that
\[
\eta E\,\varepsilon_i = O(\eta^2 E^2),
\]
so that the contribution of the local gradient term is of
second-order magnitude.
\end{assumption}

Under this assumption, the local term in Eq. \eqref{eq:E_final_corrected} satisfies
\[
\eta E\,
\nabla_{\theta}\mathcal{L}(\theta_i^{t-1};\mathcal{D}_i)
= O(\eta^2 E^2),
\]
and is absorbed into the second-order error.

Substituting into Eq. \eqref{eq:E_final_corrected} gives
\begin{equation}
\boxed{
\theta_g^t - \theta_i^{t-1}
=
-\eta E\,
\nabla_{\theta}\mathcal{L}(\theta_i^{t-1};\mathcal{D})
+
O(\eta^2 E^2).
}
\label{eq:F1_final_smallgrad}
\end{equation}




Hence,
\begin{equation}
\nabla_{\theta}\mathcal{L}(\theta_i^{t-1};\mathcal{D})
\;\approx\;
\frac{1}{\eta E}\bigl(\theta_i^{t-1} - \theta_g^{t}\bigr).
\end{equation}

Since $\eta$ and $E$ are fixed hyperparameters during training, the 
factor $1/(\eta E)$ acts as a constant scalar.
Because the importance score $I_O$ depends only on squared magnitudes 
and is used solely for \emph{relative} ranking across parameters, 
multiplying the gradient estimate by a positive constant does not 
change the behavior of the method.  
Therefore, without loss of generality, we set the overall scaling to $1$ 
and use the simplified approximation
\begin{equation}
\nabla_{\theta}\mathcal{L}(\theta_i^{t-1};\mathcal{D})
\;\approx\;
\theta_i^{t-1} - \theta_g^{t}.
\end{equation}

\texttt{FedOBP} introduces an interpretable and practical criterion by directly quantifying the discrepancy between local and global parameters for $\mathcal{D}$. This formulation enables a more principled selection of parameters for personalization, allowing clients to identify and retain only the most impactful parameters based on their relevance to global knowledge for efficient personalization. Specifically, local parameters $\theta_i^{t-1,k}$ with higher importance scores $I_{O}({\theta}_i^{t-1,k};\mathcal{D})$ are the most influential in improving global model performance. Personalizing these parameters is crucial to avoid excessive alignment with the global model, preserving strong local performance. This allows clients to personalize only a small set of important parameters while sharing the rest globally, balancing local adaptation and global collaboration.

\begin{table*}[!]
    \centering
    \caption{Statistical information of used datasets on clients.}
    \begin{tabular}{cccccc}
        \toprule
        Dataset & Samples & Classes & Description & Resolution & Year \\
        \midrule
        CIFAR10/100  & 60,000  & 10/100 & Images in 10/100 classes including airplanes, cars, birds, etc. & $32\times32$ & 2009 \\
        EMNIST  & 805,263  & 62 & Extended MNIST with letters and digits & $28\times28$ & 2017 \\
        FMNIST & 70,000 & 10 & Fashion item images & $28\times28$ & 2017 \\
        MNIST & 70000 & 10 & Handwritten digits & $28\times28$ & 1998 \\
        MEDMNISTA/C & 58,850/23,600 & 11 & Biomedical images on abdominal CT. & $28\times28$ & 2019 \\
        SVHN & 600,000 & 10 & Street view house numbers & $32\times32$ & 2011 \\
        \bottomrule
    \end{tabular}
    \label{tab:data details}
\end{table*}

\begin{table}[!]
\centering
\caption{Quantile($q$) thresholds and corresponding number of personalized parameters for $\text{Dir}(0.1)$ and $\text{Dir}(0.5)$ across four datasets.}
\centering
\begin{tabular}{c|cccc}
\toprule
\multirow{2}{*}{Datasets} & \multicolumn{2}{c}{$\text{Dir}(0.1)$} & \multicolumn{2}{c}{$\text{Dir}(0.5)$} \\ 
& $q$ & Number & $q$ & Number \\ \hline
\multicolumn{1}{c|}{CIFAR10} & 0.99998 & 18 & 0.99979 & 185 \\    
\multicolumn{1}{c|}{CIFAR100} & 0.9998 & 185 & 0.99992 & 74 \\     
\multicolumn{1}{c|}{EMNIST} & 0.995 & 3,044 & 0.9991 & 548 \\        
\multicolumn{1}{c|}{SVHN} & 0.99997 & 27 & 0.9998 & 176 \\           
\multicolumn{1}{c|}{MNIST} & 0.9998 & 117 & 0.99993 & 41 \\  
\multicolumn{1}{c|}{FMNIST} & 0.99993 & 41 & 0.9999 & 59 \\          
\multicolumn{1}{c|}{MEDMNISTA} & 0.9999 & 53 & 0.99995 & 30 \\      
\multicolumn{1}{c|}{MEDMNISTC} & 0.9997 & 175 & 0.99995 & 30 \\       
\bottomrule
\end{tabular}
\label{quantile}
\end{table}

\section{Experiments}
\label{exp:settings}
\subsection{Experimental Setup}
\textbf{Datasets}.
We evaluate the proposed method on four benchmark datasets: EMNIST, CIFAR-10, CIFAR-100, and SVHN, covering handwritten character, object, and digit classification tasks. Detailed dataset statistics are reported in Table~\ref{tab:data details}. For each dataset, data are evenly divided into non-overlapping training and test sets across clients. To simulate non-IID data heterogeneity, we partition each dataset into 100 subsets and allocate them to 100 clients according to a Dirichlet distribution $\text{Dir}(\alpha)$ with $\alpha \in \{0.1, 0.5\}$, where a smaller $\alpha$ indicates greater heterogeneity.
The federated learning setup uses $T=400$ global communication rounds, 100 clients, and a participation rate of $\gamma=0.1$. Local training is performed with SGD using a learning rate of $\eta=0.01$, batch size 32, and 5 local epochs. Results are averaged over four independent runs, with both mean and standard deviation reported.

\textbf{Baselines}. We implement the baselines based on an open-source \href{https://github.com/KarhouTam/FL-bench}{benchmark}.
We compare the performance of our \texttt{FedOBP} algorithm with eight current PFL methods, including 
FedPer~\citep{arivazhagan2019federated},
APFL~\citep{deng2020adaptive},
LG-FedAvg~\citep{liang2020think},
FedRep~\citep{collins2021exploiting},
pFedFDA~\citep{mclaughlinpersonalized},
FLUTE~\citep{liu2024federated},
FedDPA~\citep{yang2023dynamic}, 
FLOCO~\citep{grinwaldfederated} and FedALA~\citep{zhang2023fedala}.
FedAvg~\citep{mcmahan2017communication} and Local-Only served as baselines for assessing generalization and personalization performance.

\textbf{Model and Hyperparameters}. We adopt a simple four-layer CNN as the backbone model, consisting of two convolutional layers and two fully connected layers, with the final fully connected layer serving as the classifier. To normalize parameter importance, we consider two strategies, LayerNorm and GlobalNorm. LayerNorm performs layer-wise min--max normalization following~\citep{yang2023dynamic}, whereas GlobalNorm applies min--max normalization across all parameters. These two strategies, together with the variant without normalization, are compared in the ablation study in Section~\ref{app:ablation}. The evaluation metric is the average accuracy, computed by first evaluating each client \(i \in \mathcal{C}\) on its local dataset \(\mathcal{D}_i\) and then averaging across all clients. Table~\ref{quantile} reports the \texttt{FedOBP} quantile settings for eight datasets under \(\alpha \in \{0.1, 0.5\}\).

\begin{table*}[!]
\footnotesize
\centering
\caption{Average (standard deviation) test accuracy (\%) on four datasets. \textbf{Bold} and \underline{underlined} indicate the best and second-best respectively.}
\vspace{-1em}
\begin{tabular}
{>{\centering\arraybackslash}p{1.2cm}|>{\centering\arraybackslash}p{1.5cm}|>{\centering\arraybackslash}p{1.5cm}|>{\centering\arraybackslash}p{1.5cm}|>{\centering\arraybackslash}p{1.5cm}|>{\centering\arraybackslash}p{1.5cm}|>{\centering\arraybackslash}p{1.5cm}|>{\centering\arraybackslash}p{1.5cm}|>{\centering\arraybackslash}p{1.5cm}}
\toprule
\multicolumn{1}{c|}{Dataset} & \multicolumn{2}{c|}{CIFAR10} & \multicolumn{2}{c|}{CIFAR100} & \multicolumn{2}{c|}{EMNIST} & \multicolumn{2}{c}{SVHN} 
\\
\multicolumn{1}{c|}{Partition} & \multicolumn{1}{c}{$\text{Dir}(0.1)$} & \multicolumn{1}{c|}{Dir(0.5)} & \multicolumn{1}{c}{$\text{Dir}(0.1)$} & \multicolumn{1}{c|}{Dir(0.5)} & \multicolumn{1}{c}{$\text{Dir}(0.1)$} & \multicolumn{1}{c|}{Dir(0.5)} & \multicolumn{1}{c}{$\text{Dir}(0.1)$} & \multicolumn{1}{c}{Dir(0.5)}  
\\ 
\midrule
\multicolumn{1}{c|}{Local-Only} & 80.87(0.12) & 54.78(0.24) & 37.88(0.50) & 15.76(0.39) & 92.72(0.81) & 85.02(0.23) & 90.15(0.26) & 77.71(0.61) 
\\
\multicolumn{1}{c|}{FedAvg} & 58.15(0.26) & 63.98(0.29) & 26.18(0.36) & 25.40(0.22) & 82.11(0.12) & 84.02(0.17) & 88.21(0.76) & 90.24(0.66)  \\
\midrule
\multicolumn{1}{c|}{FedPer} & 84.98(0.43) & 65.87(0.67) & \underline{42.09(0.47)} & 20.68(0.27) & 94.20(0.02) & \underline{87.55(0.09)} & \underline{94.73(0.23)} & 89.46(0.65)  \\
\multicolumn{1}{c|}{APFL} & 59.37(1.23) & 63.78(0.21) & 26.55(0.47) & 25.10(0.06) & 81.87(0.16) & 83.85(0.15) & 89.51(0.74) & \underline{91.05(0.41)}  \\
\multicolumn{1}{c|}{LG-FedAvg} & 81.74(0.25) & 57.46(0.87) & 39.08(0.73) & 16.89(0.73) & 93.69(0.09) & 86.32(0.34) & 91.68(0.41) & 81.63(1.10)  \\
\multicolumn{1}{c|}{FedRep} & 84.56(0.26) & 63.63(0.49) & 39.35(0.35) & 16.83(0.18) & \underline{94.36(0.22)} & 87.38(0.56) & 94.16(0.22) & 86.91(0.43)  \\
\multicolumn{1}{c|}{pFedFDA} & \underline{86.43(0.10)} & \underline{68.72(0.19)} & 41.72(0.45) & 16.71(0.77) & 93.39(0.13) & 86.24(0.21) & 93.95(0.14) & 87.53(0.28)  \\
\multicolumn{1}{c|}{FLUTE} & 74.78(0.00) & 48.25(1.19) & 31.61(0.00) & 12.63(1.01) & 80.32(0.00) & 63.87(0.44) & 72.24(0.00) & 43.06(0.89)  \\
\multicolumn{1}{c|}{FedDPA} & 81.39(0.02) & 57.37(1.19) & 38.88(0.11) & 16.74(1.27) & 93.73(0.02) & 86.64(0.29) & 91.99(0.01) & 82.83(1.39)  \\
\multicolumn{1}{c|}{FLOCO} & 80.77(0.73) & 66.26(0.18) & 27.98(0.63) & 18.25(0.82) & 88.74(0.06) & 85.71(0.20) & 93.37(0.51) & 89.76(0.62)  \\
\multicolumn{1}{c|}{FedALA} & 56.94(0.09) & 63.58(0.00) & 26.00(0.09) & \underline{25.69(0.00)} & 82.06(0.05) & 83.89(0.00) & 87.61(0.12) & 89.91(0.00)  \\
\multicolumn{1}{c|}{FedSelect} & 79.89(0.00) & 53.59(0.00) & 36.33(0.00) & 15.12(0.00) & 92.91(0.00) & 84.57(0.00) & 90.14(0.00) & 77.58(0.00)  \\
\midrule
\multicolumn{1}{c|}{\texttt{FedOBP}} & \textbf{87.02(0.17)} & \textbf{70.22(0.16)} & \textbf{50.37(0.16)} & \textbf{27.05(0.30)} & \textbf{94.36(0.04)} & \textbf{88.82(0.05)} & \textbf{95.60(0.03)} & \textbf{91.75(0.47)}  \\
\bottomrule
\end{tabular}
\label{accuracy}
\end{table*}

\begin{table*}[!]
\footnotesize
\centering
\caption{Average (standard deviation) test accuracy (\%) on multiple datasets. \textbf{Bold} and \underline{underlined} indicate the best and second-best respectively.}
\begin{tabular}
{>{\centering\arraybackslash}p{1.2cm}|>{\centering\arraybackslash}p{1.5cm}|>{\centering\arraybackslash}p{1.5cm}|>{\centering\arraybackslash}p{1.5cm}|>{\centering\arraybackslash}p{1.5cm}|>{\centering\arraybackslash}p{1.5cm}|>{\centering\arraybackslash}p{1.5cm}|>{\centering\arraybackslash}p{1.5cm}|>{\centering\arraybackslash}p{1.5cm}}
\toprule
\multicolumn{1}{c|}{Dataset} & \multicolumn{2}{c|}{MNIST} & \multicolumn{2}{c|}{FMNIST} & \multicolumn{2}{c|}{MEDMNISTA} & \multicolumn{2}{c}{MEDMNISTC} 
\\
\multicolumn{1}{c|}{Partition} & \multicolumn{1}{c}{$\text{Dir}(0.1)$} & \multicolumn{1}{c|}{$\text{Dir}(0.5)$} & \multicolumn{1}{c}{$\text{Dir}(0.1)$} & \multicolumn{1}{c|}{$\text{Dir}(0.5)$} & \multicolumn{1}{c}{$\text{Dir}(0.1)$} & \multicolumn{1}{c|}{$\text{Dir}(0.5)$} & \multicolumn{1}{c}{$\text{Dir}(0.1)$} & \multicolumn{1}{c}{$\text{Dir}(0.5)$}  
\\ 
\midrule
\multicolumn{1}{c|}{Local-Only} & 97.49(0.05) & 94.44(0.13) & 94.93(0.08) & 86.49(0.24) & 67.22(0.05) & 41.08(0.07) & 66.92(0.07) & 41.21(0.15)  \\
\multicolumn{1}{c|}{FedAvg} & 98.58(0.13) & 98.89(0.08) & 87.29(0.45) & 90.37(0.22) & 18.10(0.00) & 18.28(0.55) & 22.28(0.00) & 22.33(0.11)  \\ 
\midrule
\multicolumn{1}{c|}{FedPer} & \underline{99.03(0.06)} & 98.10(0.22) & \underline{96.24(0.19)} &91.06(0.38) & \underline{67.44(0.03)} & 41.10(0.05) & 66.83(0.02) & 41.35(0.06)  \\
\multicolumn{1}{c|}{APFL} & 98.98(0.18) & \textbf{99.04(0.09)} & 89.37(0.51) & 91.00(0.39) & 58.50(0.97) & 35.89(1.78) & 56.06(0.74) & 38.81(0.19)  \\
\multicolumn{1}{c|}{LG-FedAvg} & 97.82(0.13) & 95.46(0.45) & 95.24(0.07) & 87.36(0.30) & 67.23(0.07) & 41.10(0.13) & \textbf{66.97(0.01)} & 41.34(0.03)  \\
\multicolumn{1}{c|}{FedRep} & 98.59(0.10) & 97.16(0.20) & 95.80(0.20) & 89.85(0.46) & 67.09(0.12) & 41.08(0.08) & 66.35(0.16) & 41.18(0.05)  \\
\multicolumn{1}{c|}{pFedFDA} & 98.78(0.09) & 97.80(0.22) & 96.09(0.15) & 91.10(0.48) & \textbf{67.47(0.00)} & \underline{41.22(0.01)} & 66.79(0.00) & \textbf{41.45(0.00)}  \\
\multicolumn{1}{c|}{FLUTE} & 91.13(0.00) & 80.48(0.00) & 88.56(0.00) & 72.83(0.00) & \textbf{67.47(0.00)} & 41.14(0.00) & 66.79(0.00) & 41.27(0.00)  \\
\multicolumn{1}{c|}{FedDPA} & 98.04(0.00) & 95.92(0.52) & 95.06(0.00) & 88.01(0.51) & 67.02(0.10) & 40.34(0.97) & 66.66(0.05) & 40.94(0.34)  \\
\multicolumn{1}{c|}{FLOCO} & 98.80(0.17) & 98.50(0.10) & 95.06(0.58) & \underline{91.38(0.54)} & 59.47(0.23) & 33.40(0.44) & 60.66(1.26) & 34.78(0.89)  \\
\multicolumn{1}{c|}{FedALA} & 98.58(0.04) & 98.88(0.04) & 86.98(0.04) & 90.18(0.00) & 18.10(0.00) & 17.96(0.00) & 9.03(0.00) & 22.27(0.00)  \\
\multicolumn{1}{c|}{FedSelect} & 97.46(0.00) & 94.44(0.00) & 93.83(0.00) & 86.21(0.00) & 67.13(0.00) & 40.96(0.00) & 66.62(0.00) & 41.32(0.00)  \\
\midrule
\multicolumn{1}{c|}{\texttt{FedOBP}} & \textbf{99.34(0.02)} & \underline{98.97(0.04)} & \textbf{96.89(0.03)} & \textbf{93.11(0.04)} & 67.31(0.00) & \textbf{41.22(0.13)} & \underline{66.84(0.06)} & \underline{41.36(0.09)}  \\
\bottomrule
\end{tabular}
\label{accuracy_2}
\end{table*}

\subsection{Performance Results}
\label{exp:performance}
Table~\ref{accuracy} compares the accuracy of our method with eight baselines on four image classification tasks. The results show that our method consistently achieves the best performance under different levels of non-IID data heterogeneity. Under the $\text{Dir}(0.1)$ setting, \texttt{FedOBP} surpasses the second-best method by 0.59\%, 8.28\%, and 0.16\% on CIFAR10, CIFAR100, and EMNIST, respectively, and improves upon FedPer by 0.87\% on SVHN. Under the $\text{Dir}(0.5)$ setting, \texttt{FedOBP} remains superior on all datasets. These findings verify the effectiveness of \texttt{FedOBP} for personalization under non-IID data distributions.

Table~\ref{accuracy_2} summarizes the average test accuracy on MNIST, FMNIST and MEDMNISTA/C, comparing \texttt{FedOBP} with several SOTA baselines. On MNIST and FMNIST, \texttt{FedOBP} achieves the best performance under $\text{Dir}(0.1)$, surpassing the second-best methods by 0.31\% and 0.65\%, respectively. Under $\text{Dir}(0.5)$, it remains competitive on MNIST and achieves the best result on FMNIST with a 1.73\% improvement over \texttt{FLOCO}. On MEDMNISTA, \texttt{FedOBP} is 0.16\% below the best result under $\text{Dir}(0.1)$ and ties for the best performance with \texttt{pFedFDA} under $\text{Dir}(0.5)$. On MEDMNISTC, \texttt{FedOBP} also delivers competitive results under both heterogeneity settings, although it falls slightly short of the best performance.

\subsection{Convergence}
\begin{figure*}[ht]
    \centering
    \begin{minipage}[b]{1\linewidth}
        \centering
        \includegraphics[width=\linewidth]{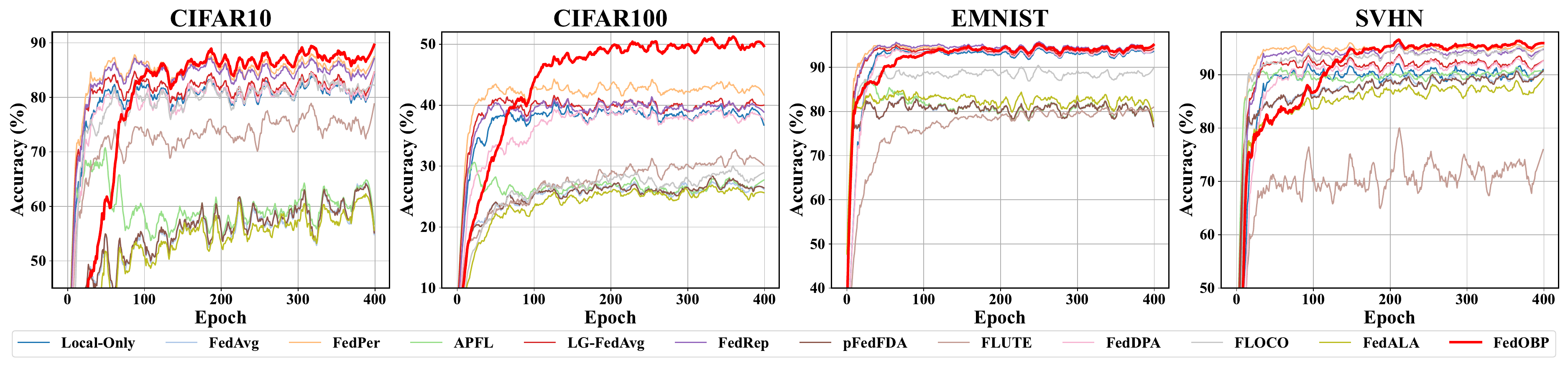}
    \end{minipage}
    
    \begin{minipage}[b]{1\linewidth}
        \centering
        \includegraphics[width=\linewidth]{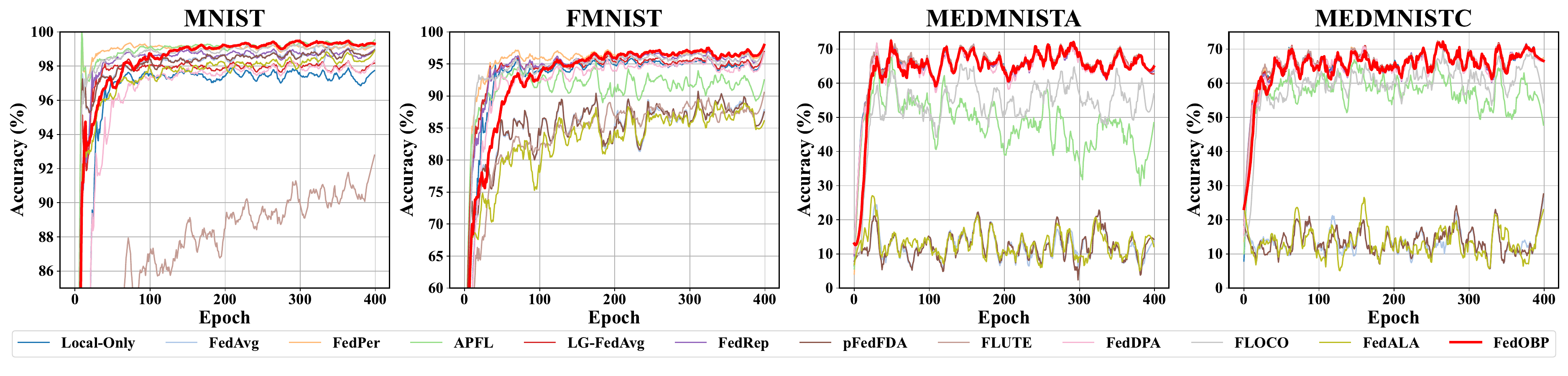}
    \end{minipage}
    
    \vspace{-0.5cm}
    \caption{Convergence comparison of {\tt FedOBP} and eleven baseline methods $\alpha=0.1$ on the 4-layer CNN model across eight datasets.}
    \vspace{-0.5em}
    \label{fig:convergence01}
\end{figure*}

\begin{figure*}[ht]
    \centering
    \begin{minipage}[b]{1\linewidth}
        \centering
        \includegraphics[width=\linewidth]{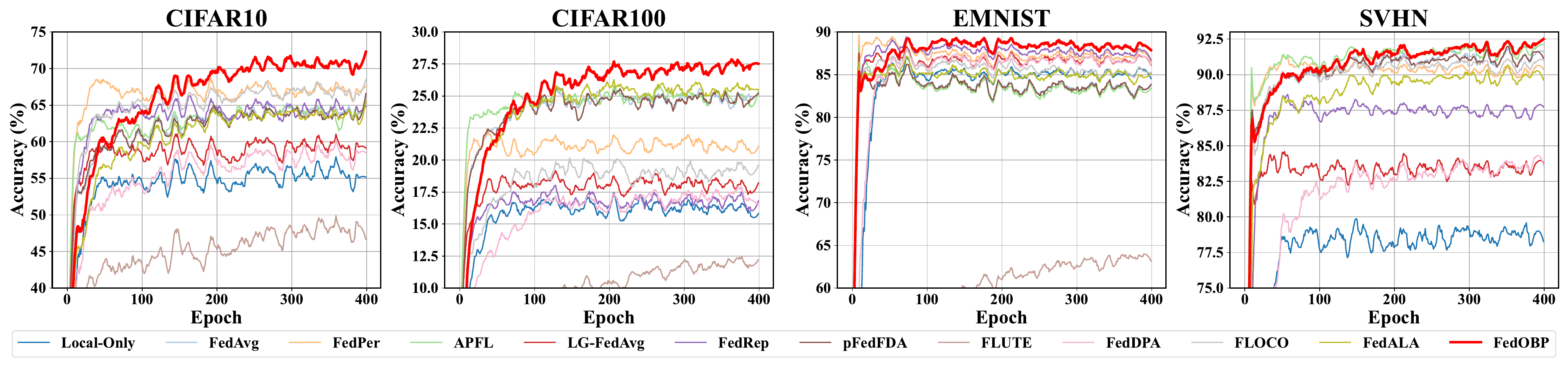}
    \end{minipage}
    
    \begin{minipage}[b]{1\linewidth}
        \centering
        \includegraphics[width=\linewidth]{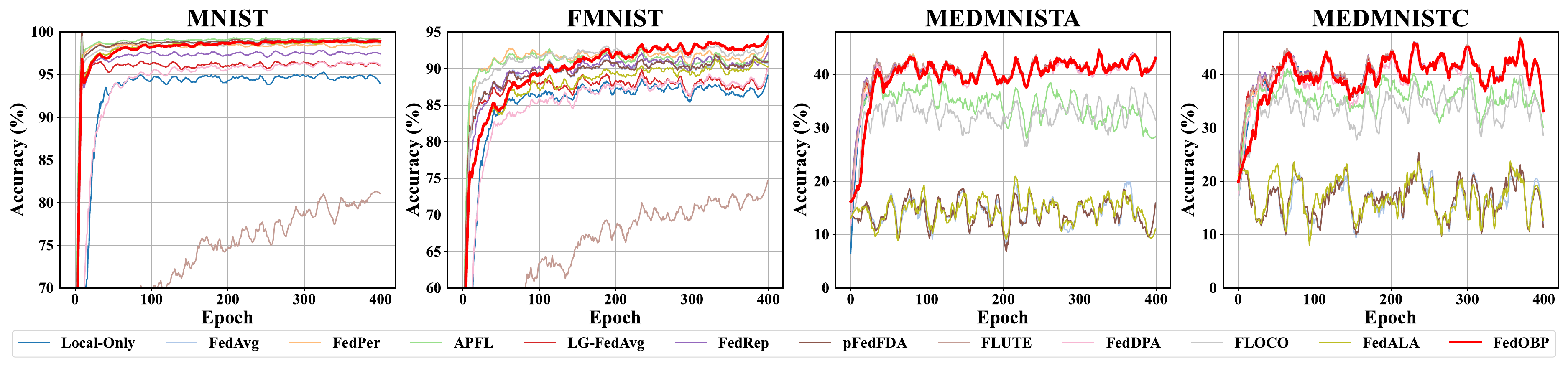}
    \end{minipage}
    
    \vspace{-0.5cm}
    \caption{Convergence comparison of {\tt FedOBP} and eleven baseline methods with $\alpha=0.5$ on the 4-layer CNN model across eight datasets.}
    \vspace{-0.5em}
    \label{fig:convergence05}
\end{figure*}

Figures~\ref{fig:convergence01} and~\ref{fig:convergence05} show the convergence results under two levels of data heterogeneity. Under $\alpha=0.1$, \texttt{FedOBP} demonstrates strong convergence performance across all datasets. On CIFAR10, it achieves the highest accuracy with competitive convergence. pFedFDA and FedPer also perform well, both exceeding 85\%, whereas APFL, FLUTE, and FedALA converge more slowly and remain below 75\%. On CIFAR100 and EMNIST, \texttt{FedOBP} achieves both the fastest convergence and the best final accuracy. On MEDMNISTA and MEDMNISTC, although its final accuracy is slightly lower than that of the best-performing methods, it converges faster than most baselines. Among the model-decoupling methods, LG-FedAvg and pFedFDA are also competitive, surpassing 67\% and 66\% accuracy on MEDMNISTA and MEDMNISTC, respectively.

Under $\alpha=0.5$, \texttt{FedOBP} continues to perform competitively across all datasets. On CIFAR10 and CIFAR100, it achieves the best accuracy with competitive convergence and reaches near-optimal performance at around 200 epochs. It also ranks first on SVHN with the fastest convergence, while maintaining rapid convergence on EMNIST. Similar patterns are observed on MNIST, FMNIST, MEDMNISTA, and MEDMNISTC. Overall, these results verify that \texttt{FedOBP} achieves a strong trade-off between convergence speed and final accuracy under different levels of data heterogeneity.

\subsection{Importance Scores}
\label{Importance_Scores}
\begin{figure*}[!]
    \centering
    \includegraphics[width=\linewidth]{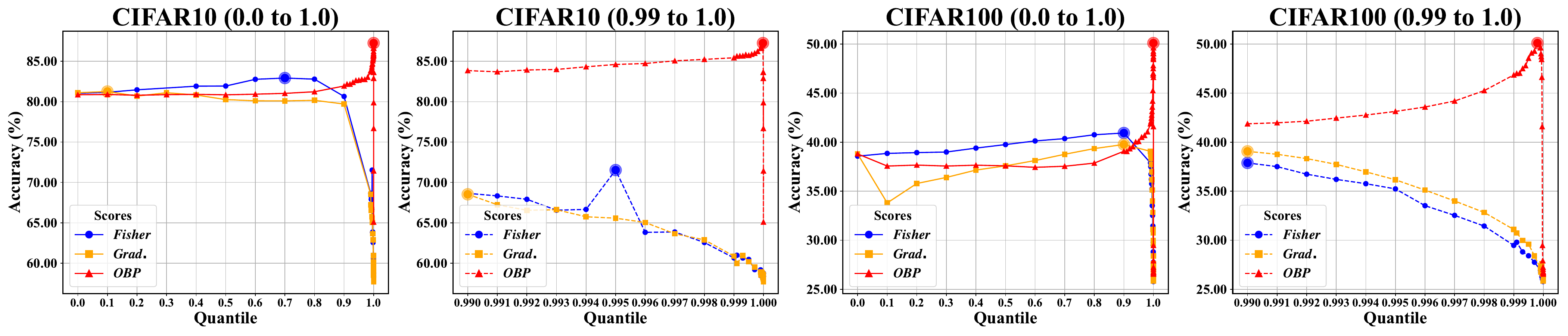}
    \includegraphics[width=\linewidth]{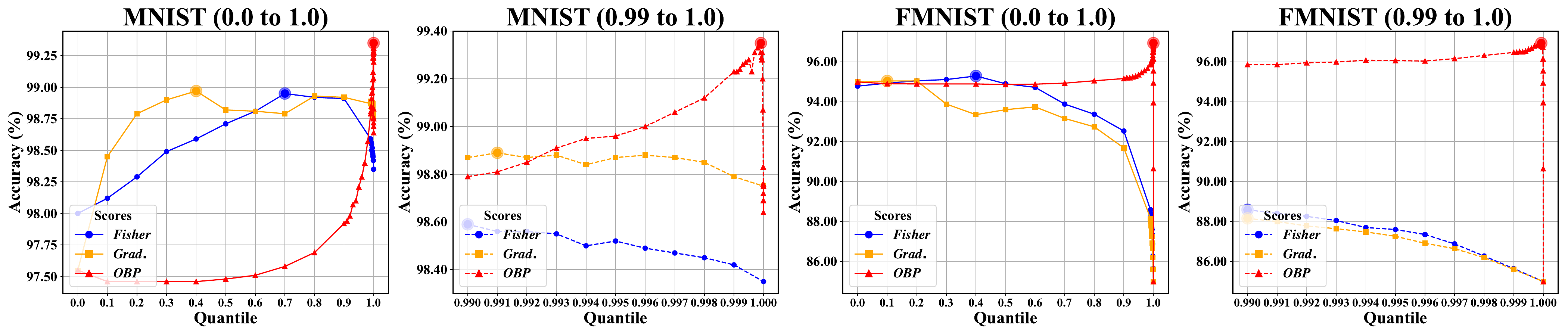}
    \includegraphics[width=\linewidth]{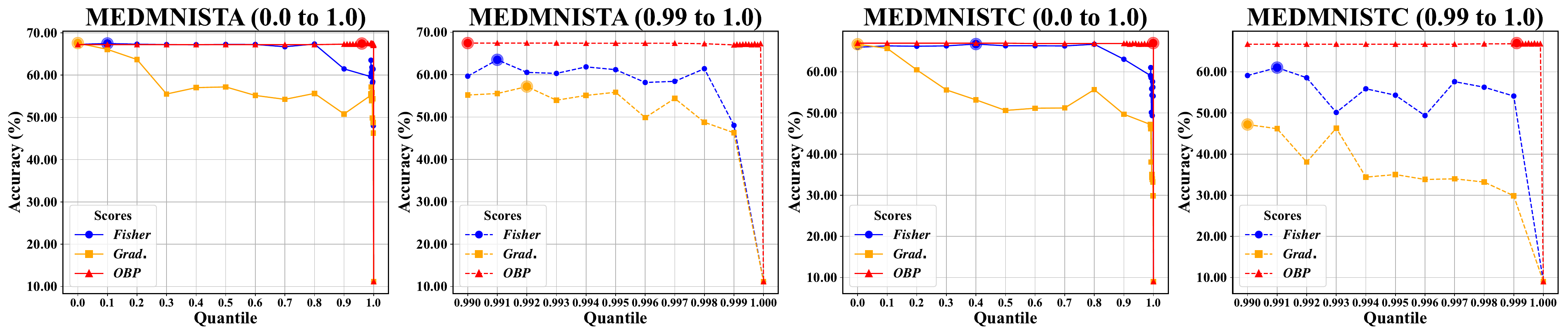}
    \vspace{-0.6cm}
    \caption{Comparison of three scores (Gradient $I_{G}(\cdot)$, Fisher $I_{F}(\cdot)$, and OBP $I_{O}(\cdot)$) across six datasets under different quantile settings.}
    \vspace{-0.4cm}
    \label{fig:score}
\end{figure*}

We further analyze the performance of three types of scores (Gradient \( I_{G}(\cdot) \), Fisher \( I_{F}(\cdot) \), and \texttt{FedOBP} \( I_{O}(\cdot) \)) with different quantile \( q \) settings ranging from $0.0$ to $1.0$, where a larger quantile indicates a smaller proportion ($1-q$) of personalized parameters. 

As shown in Figure~\ref{fig:score}, across all datasets, as the quantile \( q \) increases, the number of personalized parameters \( \bm{u}_i \) decreases while the number of global parameters \( \bm{v}_i \) increases. Correspondingly, the performance of all three scores generally exhibits a consistent trend, namely, it first improves and then declines. This behavior indicates a clear phase transition, where each score reaches its peak performance at a specific quantile value depending on the dataset. On CIFAR10, \( I_{G}(\cdot) \) reaches its best performance at \( q=0.1 \), \( I_{F}(\cdot) \) peaks at \( q=0.7 \), and \( I_{O}(\cdot) \) achieves its optimum at \( q=0.9999 \). These quantile values correspond to approximately \( 790{,}684 \), \( 263{,}561 \), and \( 87 \) personalized parameters, respectively. This result shows that \( I_{O}(\cdot) \) requires substantially fewer personalized parameters to achieve optimal performance. Moreover, when the number of FL rounds increases to \( 400 \), the number of personalized parameters required by \( I_{O}(\cdot) \) further decreases to \( 18 \), as reported in Table~\ref{quantile}. On CIFAR100, both \( I_{G}(\cdot) \) and \( I_{F}(\cdot) \) attain their best performance at \( q=0.9 \), corresponding to \( 87{,}853 \) personalized parameters, whereas \( I_{O}(\cdot) \) peaks at \( q=0.9998 \), requiring only \( 185 \) personalized parameters. 

The similar phenomenon can also be observed on MNIST, FMNIST and MEDMNISTA/C. On MNIST, to achieve optimal performance, \( I_{G}(\cdot) \) requires approximately \( 60\% \) personalized parameters, \( I_{F}(\cdot) \) requires about \( 30\% \), whereas \( I_{O}(\cdot) \) requires less than \( 0.1\% \). On FMNIST, \( I_{G}(\cdot) \) peaks at \( q=0.1 \), \( I_{F}(\cdot) \) reaches its maximum at \( q=0.4 \), and \( I_{O}(\cdot) \) peaks at \( q=0.99993 \), again indicating that \( I_{O}(\cdot) \) can identify a much smaller set of necessary personalized parameters. For MEDMNISTA and MEDMNISTC, the OBP-based score exhibits more stable performance across different quantile ranges than the gradient-based and Fisher-based scores. Overall, these results consistently show that \( I_{O}(\cdot) \) can more accurately identify the truly necessary personalized parameters, achieving strong performance with only a small personalization budget.

\subsection{Ablation Studies}
\label{app:ablation}
The ablation study compares three normalization strategies for the \texttt{FedOBP} score \(I_{O}(\cdot)\), namely NoNorm, LayerNorm, and GlobalNorm, across eight datasets. NoNorm uses raw \(I_{O}(\cdot)\), LayerNorm applies layer-wise normalization following~\citep{yang2023dynamic}, and GlobalNorm normalizes scores over the entire model. For GlobalNorm, we also consider a CLS variant that restricts personalized parameter selection to the classifier layer.

\begin{table}[ht]
\centering
\caption{Ablation experiment comparing NoNorm, LayerNorm, and GlobalNorm on four datasets. Additionally, the ablation study examines GlobalNorm w/o CLS and with CLS.}
\begin{tabular}{c|cccccc}
\toprule
\multirow{2}{*}{Dataset} & \multicolumn{1}{c}{\multirow{2}{*}{NoNorm}} & \multicolumn{1}{c}{\multirow{2}{*}{LayerNorm}} & \multicolumn{2}{c}{GlobalNorm} \\  
& \multicolumn{1}{c}{} & \multicolumn{1}{c}{} & w/o CLS & with CLS  
\\ 
\midrule
CIFAR10 & \underline{87.36} & 85.00 & \underline{87.36} & \textbf{87.37} \\ 
CIFAR100 & \textbf{45.98} & 41.83 & \textbf{45.98} & \underline{44.89} \\ 
EMNIST & \textbf{94.92} & 94.14 & \textbf{94.92} & \underline{94.78} \\ 
SVHN & \textbf{95.88} & \underline{93.34} & \textbf{95.88} & \textbf{95.88} \\ 
MNIST & \underline{99.38} & 99.27 & \underline{99.38} & \textbf{99.39} 
\\
FMNIST & \textbf{96.82} & 96.31 & \textbf{96.82} & \underline{96.81} 
\\
MEDMNISTA & \textbf{67.47} & \underline{67.39} & \textbf{67.47} & \textbf{67.47} 
\\
MEDMNISTC & \textbf{66.60} & 65.17 & \textbf{66.60} & \underline{66.38} 
\\ 
\bottomrule
\end{tabular}
\label{ablation2}
\end{table}

\begin{figure*}[!]
    \centering
    \includegraphics[width=\linewidth]{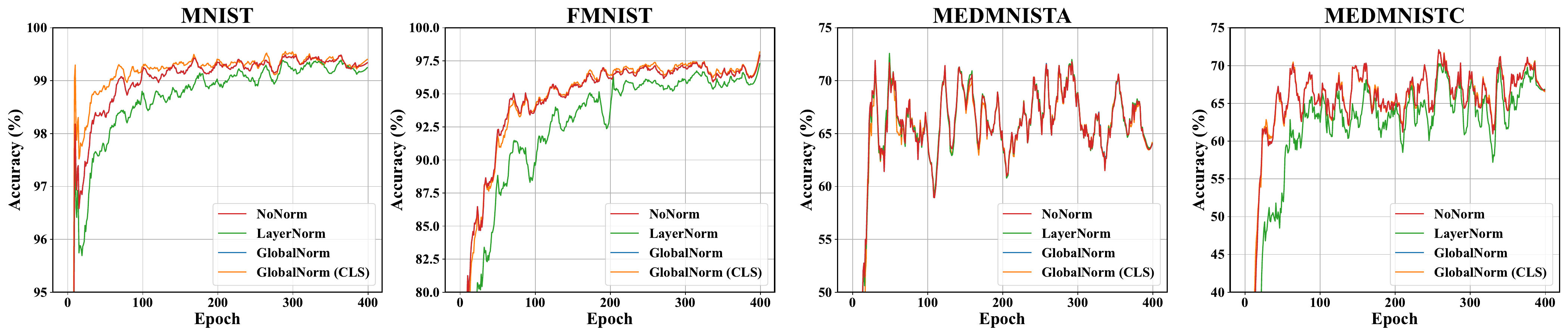}
    \includegraphics[width=\linewidth]{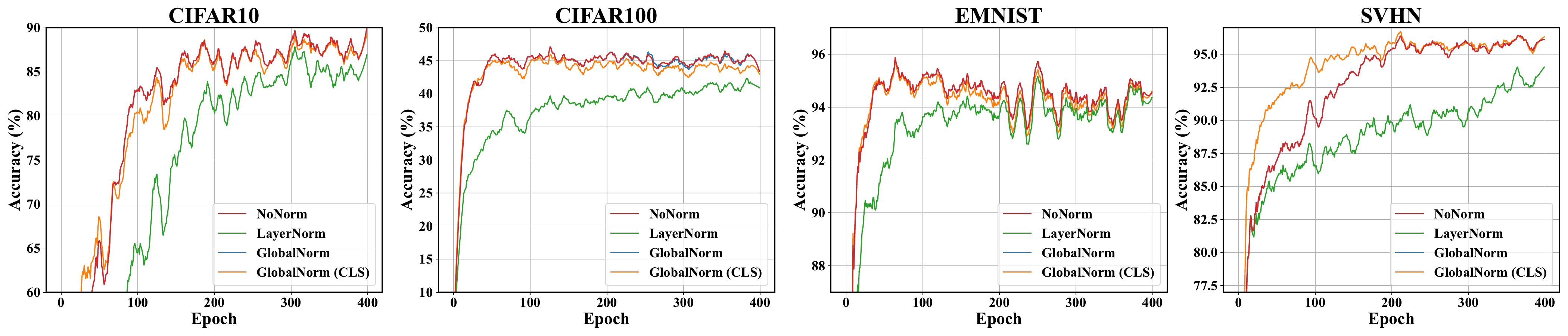}
    \caption{Ablation study of normalization strategies across eight datasets. We compare NoNorm, LayerNorm, and GlobalNorm, where GlobalNorm includes two variants, with CLS and without CLS (w/o CLS).}
    \vspace{-0.3cm}
    \label{fig:ablation}
\end{figure*}

Table~\ref{ablation2} shows that NoNorm achieves the best performance on FMNIST (96.82\%) and MEDMNISTC (66.60\%), while GlobalNorm without CLS performs best on MNIST (99.40\%) and MEDMNISTA (67.47\%). By contrast, LayerNorm consistently underperforms, indicating that global normalization is more suitable for the \texttt{FedOBP} score than layer-wise normalization. Figure~\ref{fig:ablation} shows a consistent convergence trend across all eight datasets. NoNorm and GlobalNorm achieve higher accuracy with faster convergence, whereas LayerNorm performs worse overall. For example, on CIFAR10, NoNorm and GlobalNorm without CLS both achieve 87.52\%, compared with 85.00\% for LayerNorm. On CIFAR100, they reach 45.98\%, while LayerNorm drops to 41.83\%. On EMNIST, NoNorm and GlobalNorm achieve 94.92\%, exceeding LayerNorm at 94.14\%. Similar observations hold for SVHN, MNIST, FMNIST and MEDMNISTA/C.
Restricting personalized parameter selection to the classifier layer yields comparable results on CIFAR10 and SVHN, but slightly lower performance on CIFAR100 (44.89\%) and EMNIST (94.78\%). 

\subsection{Personalized Parameter Distribution}
\begin{figure*}[!]
    \centering
    \includegraphics[width=\linewidth]{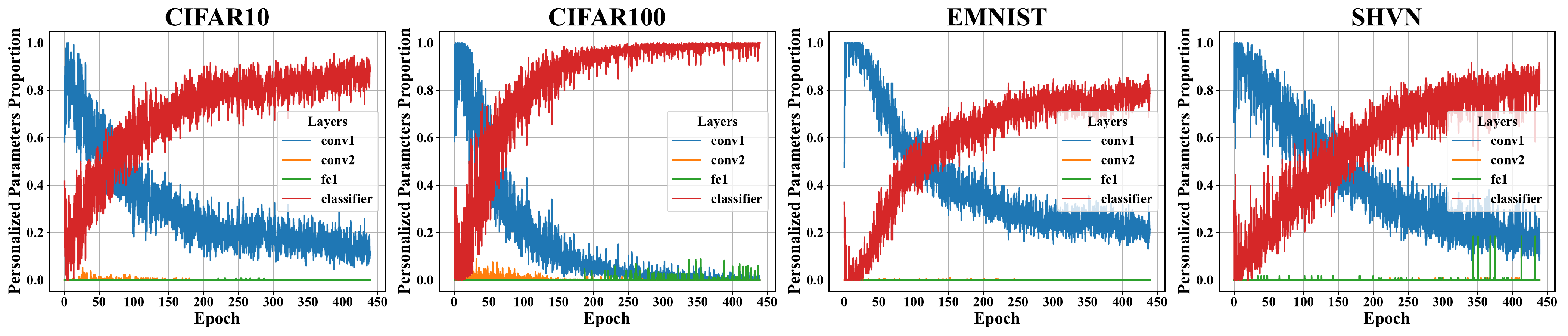}
    \includegraphics[width=\linewidth]{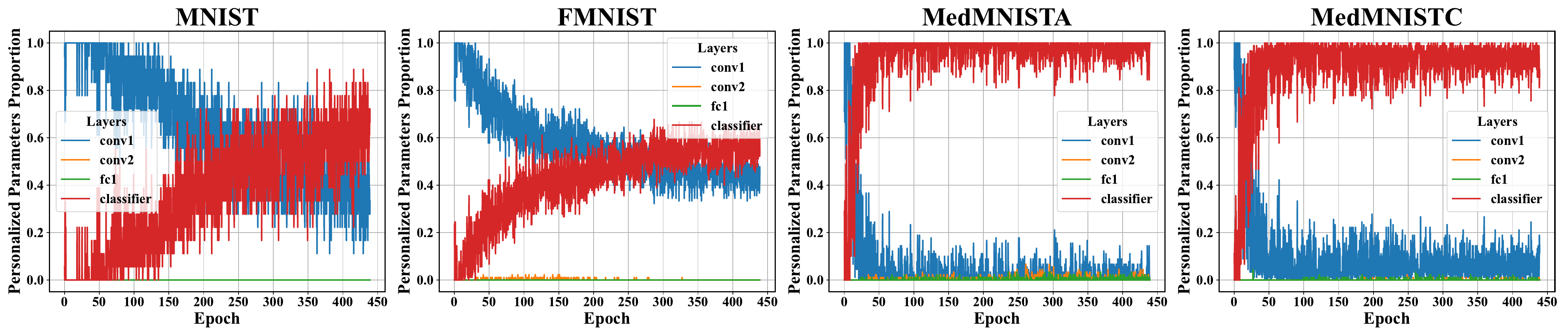}
    \vspace{-0.6cm}
    \caption{Distribution of personalized parameters across layers over FL epochs using the 4-layer CNN model on eight datasets.}
    \vspace{-0.3cm}
    \label{fig:distribution}
\end{figure*}

Figure~\ref{fig:distribution} shows the layer-wise distribution of personalized parameters in the 4-layer CNN over 450 communication epochs across eight datasets. Across all datasets, the proportion of personalized parameters in \textit{conv1} gradually decreases, while that in the classifier layer increases, indicating that personalization progressively concentrates on the classifier layer during training.
On CIFAR10 and CIFAR100, the distribution stabilizes between 300 and 450 epochs, with the classifier layer accounting for 0.9--1.0 of the personalized parameters and \textit{conv1} decreasing to 0.0--0.1. On EMNIST and SVHN, stable distributions are also observed after roughly 250--300 epochs, where the classifier layer reaches 0.7--0.8 on EMNIST and 0.7--0.9 on SVHN, while \textit{conv1} remains at 0.2--0.3 and 0.1--0.3, respectively. Similar trends are observed on MNIST, FMNIST, MEDMNISTA, and MEDMNISTC. In particular, MNIST stabilizes between 350 and 450 epochs with classifier proportions of 0.4--0.8 and \textit{conv1} proportions of 0.2--0.6, FMNIST stabilizes around 250--450 epochs with classifier and \textit{conv1} proportions of 0.5--0.6 and 0.4--0.5, respectively, and MEDMNISTA and MEDMNISTC stabilize much earlier, from 50 epochs onward, with the classifier layer dominating at 0.8--1.0 and \textit{conv1} decreasing to 0.0--0.2.

These findings are broadly consistent with prior layer-wise decoupling methods~\citep{collins2021exploiting,ohfedbabu,mclaughlinpersonalized}, which regard the final classifier layer as the primary personalization layer. In contrast, \texttt{FedOBP} identifies only a very small subset of parameters, mainly from the classifier layer, that is sufficient to achieve effective personalization and strong performance.


\subsection{Communication Overhead}
\begin{figure}[ht]
    \centering
    \begin{minipage}[b]{1\linewidth}
        \centering
        \includegraphics[width=\linewidth]{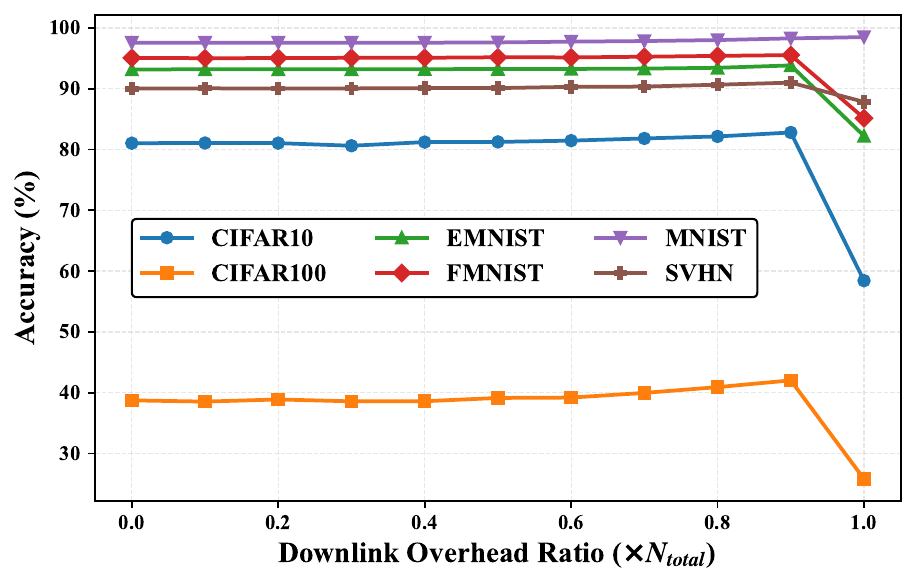}
    \end{minipage}
    \vspace{-0.7cm}
    \caption{Accuracy versus Downlink Overhead Ratio on different datasets.}
     \vspace{-0.3cm}
    \label{fig:communication}
\end{figure}
{\tt FedOBP} reduces the computational burden on resource-constrained clients by offloading the OBD-based metric computation to the server. Because this evaluation requires both the uploaded local models and the aggregated global model, the uplink communication cost in each round is fixed at $N_{total}$, i.e., the full model size. As a result, the communication savings of {\tt FedOBP} mainly arise in the downlink stage. Figure~\ref{fig:communication} reports the model accuracy under different downlink communication overhead ratios.

The results show that accuracy drops noticeably on most datasets when the downlink overhead ratio reaches $1.0$, where the full global model is transmitted without retaining personalized parameters. This suggests that relying entirely on the global model weakens adaptation to heterogeneous client data distributions. In contrast, when the downlink ratio ranges from $0$ to $0.9$, retaining a portion of personalized parameters enables the model to better fit local data distributions and consistently achieve higher accuracy than at a ratio of $1.0$. Moreover, the accuracy curves remain relatively stable over this range, indicating that reducing the downlink communication overhead causes little performance degradation as long as some personalized parameters are preserved.

\section{Conclusion}
\label{main:conclusions}
In this paper, we address the challenge in PFL of identifying which parameters should be personalized to effectively handle data heterogeneity across clients.
We propose a parameter decoupling algorithm that incorporates a quantile-based thresholding mechanism. 
In addition, we introduce an element-wise importance score, \texttt{FedOBP}, which is building on OBD pruning theory and utilizes a federated approximation of the first order derivative in the Taylor series expansion to assess the significance of each local parameter.
To accommodate the limited computational resources of edge clients, the importance metric is computed on the server side, and only the shared subset of parameters after decoupling is transmitted to clients, thereby reducing the downlink communication overhead.
Finally, we evaluate \texttt{FedOBP} on various datasets with different heterogeneity settings and show that it outperforms baselines.
Future research directions include developing adaptive thresholding methods that go beyond static quantile-based approaches with fixed thresholds. 
Furthermore, as complex architectures like ResNet require batch normalization which is known to underperform in non-IID settings, evaluating \texttt{FedOBP} with more advanced architectures remains an important direction. 
Additionally, exploring soft combinations of local and global model parameters represents another promising direction.

\bibliographystyle{IEEEtran}
\bibliography{myreference}{}

\ifCLASSOPTIONcaptionsoff
  \newpage
\fi
\vspace{-3em}

\begin{IEEEbiography}[{\includegraphics[width=0.9in,height=1.2in,clip,keepaspectratio]{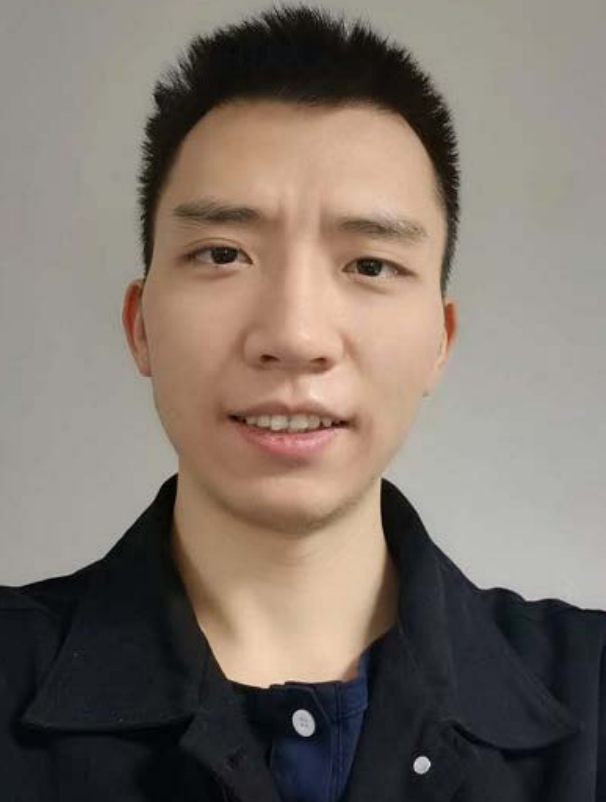}}]{Xingyan Chen} received the Ph.D. degree in computer technology from Beijing University of Posts and Telecommunications (BUPT), Beijing, China, in 2021. He is currently an Associate Professor with the School of Intelligent Engineering and Automation, BUPT. He has published in journals and conferences including \textsc{IEEE Transactions on Mobile Computing} and \textsc{IEEE INFOCOM}. His research interests include federated learning, multi-agent reinforcement learning, and stochastic optimization.
\end{IEEEbiography}
\vspace{-3em}

\begin{IEEEbiography}
[{\includegraphics[width=0.9in,height=1.2in,clip,keepaspectratio]{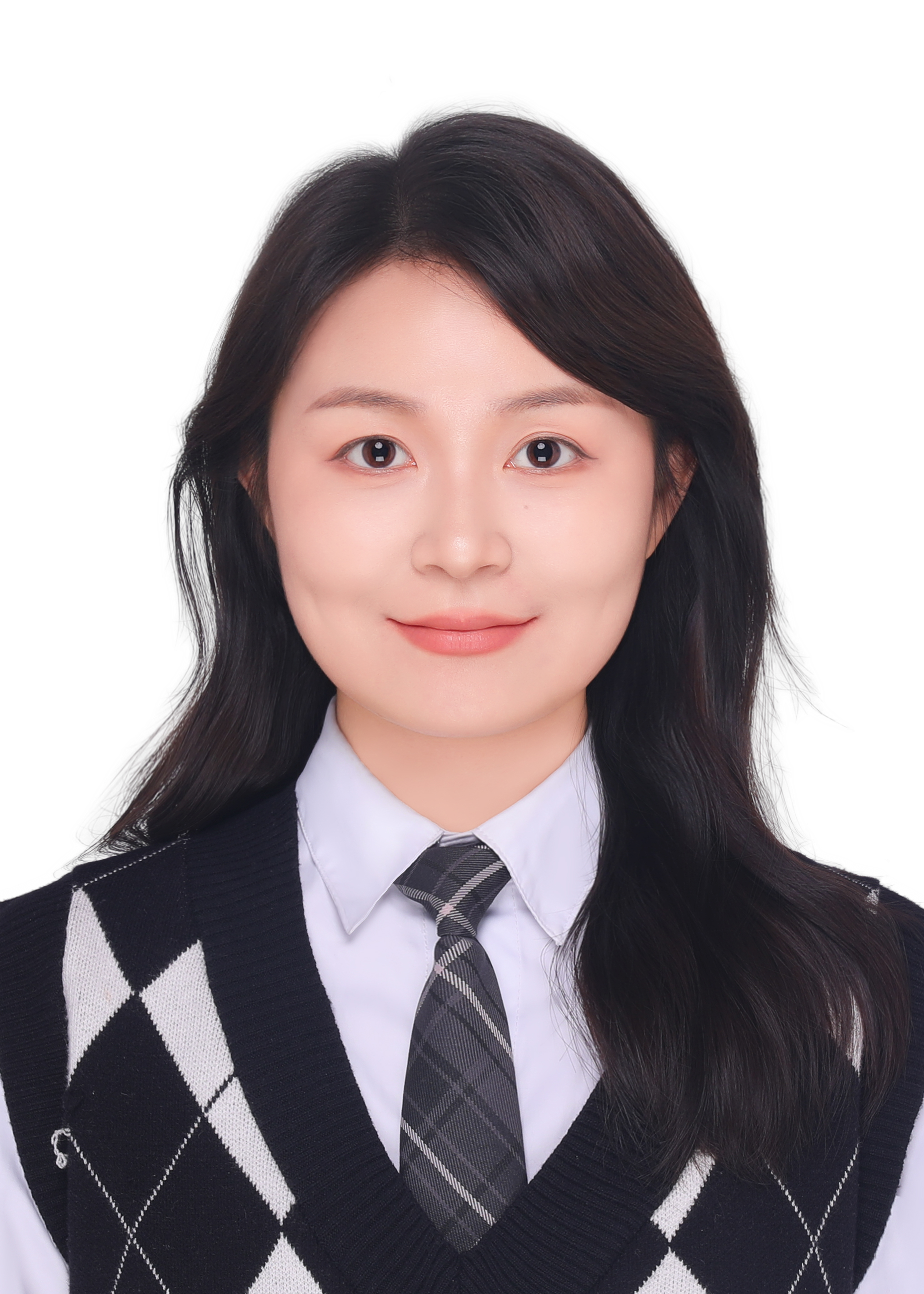}}]{Tian Du} is currently pursuing a Ph.D. in Management with a focus on Big Data Management at Southwestern University of Finance and Economics. From 2025 to 2026, she will study at Nanyang Technological University, Singapore, as a government-sponsored visiting student. Her research interests include federated learning, data privacy protection and reinforcement learning. Passionate about advancing cutting-edge data management methodologies and technologies, she is dedicated to tackling complex challenges in big data.
\end{IEEEbiography}

\vspace{-3em}

\begin{IEEEbiography}[{\includegraphics[width=1in,height=1.2in,clip,keepaspectratio]{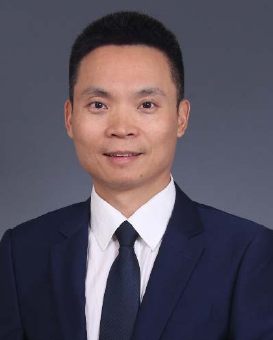}}]{Changqiao Xu} [SM'15] received the Ph.D. degree from the Institute of Software, Chinese Academy of Sciences (ISCAS), in January 2009. Since December 2009, he has been with Beijing University of Posts and Telecommunications (BUPT), Beijing, China, where he is currently a Professor with the State Key Laboratory of Networking and Switching Technology and Director of the Network Architecture Research Center. His research interests include Future Internet, mobile networking, multimedia communications, and network security. He is currently the Editor-in-Chief of \textsc{Transactions on Emerging Telecommunications Technologies (Wiley)} and a Senior Member of IEEE.
\end{IEEEbiography}

\vspace{-3em}

\begin{IEEEbiography}[{\includegraphics[width=0.9in,height=1.2in,clip,keepaspectratio]{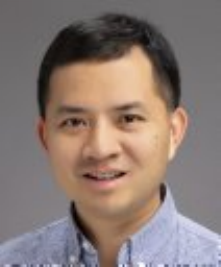}}]{Fuzhen Zhuang} received the PhD degree in
computer science from the Institute of Computing Technology, Chinese Academy of Sciences.
He is a professor at Beihang University. His research interests include transfer learning, machine learning, data mining. He has published
more than 100 papers including Nature Communications, KDD, WWW, AAAI, IEEE TKDE, IEEE
T-CYB, ACM TIST, etc.
\end{IEEEbiography}

\vspace{-3em}

\begin{IEEEbiography}[{\includegraphics[width=1in,height=1.2in,clip,keepaspectratio]{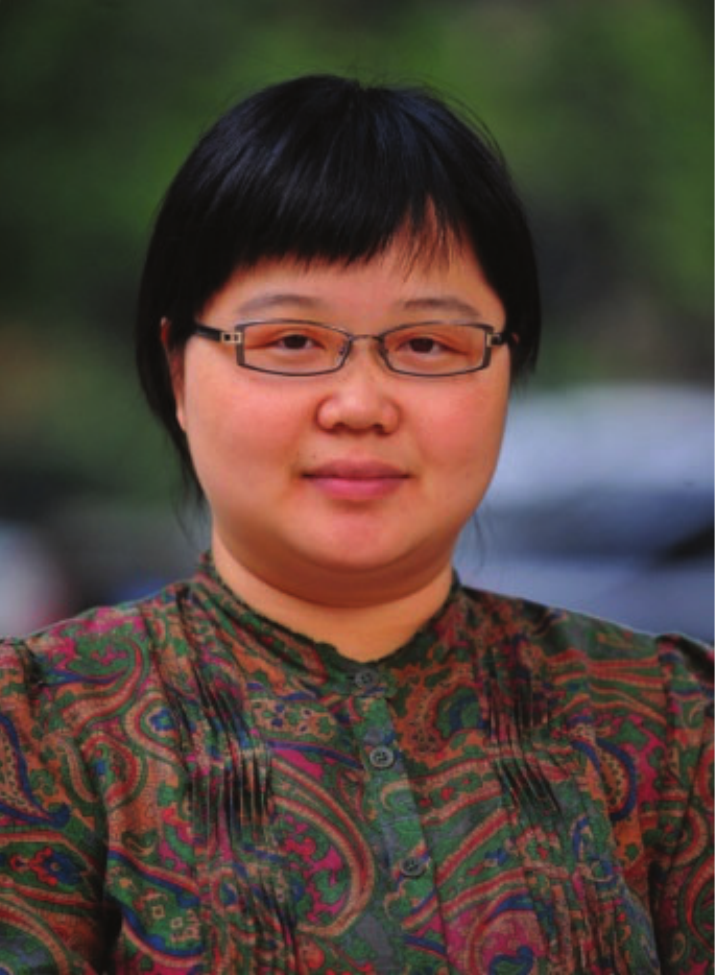}}]{Lujie Zhong} received the Ph.D. degree from the Institute of Computing Technology, Chinese Academy of Sciences, Beijing, China, in 2013. She is currently an Associate Professor with the Information Engineering College, Capital Normal University, Beijing, China. She has published papers in prestigious international journals and conferences, including \textsc{IEEE Communication Magazine}, \textsc{IEEE Transactions on Mobile Computing}, \textsc{IEEE Transactions on Multimedia}, \textsc{IEEE Internet Things Journal}, \textsc{IEEE INFOCOM} and \textsc{Acm Multimedia}, etc. Her research interests include communication networks, computer system and architecture, and mobile Internet technology.
\end{IEEEbiography}

\vspace{-3em}

\begin{IEEEbiography}[{\includegraphics[width=1in,height=1.2in,clip,keepaspectratio]{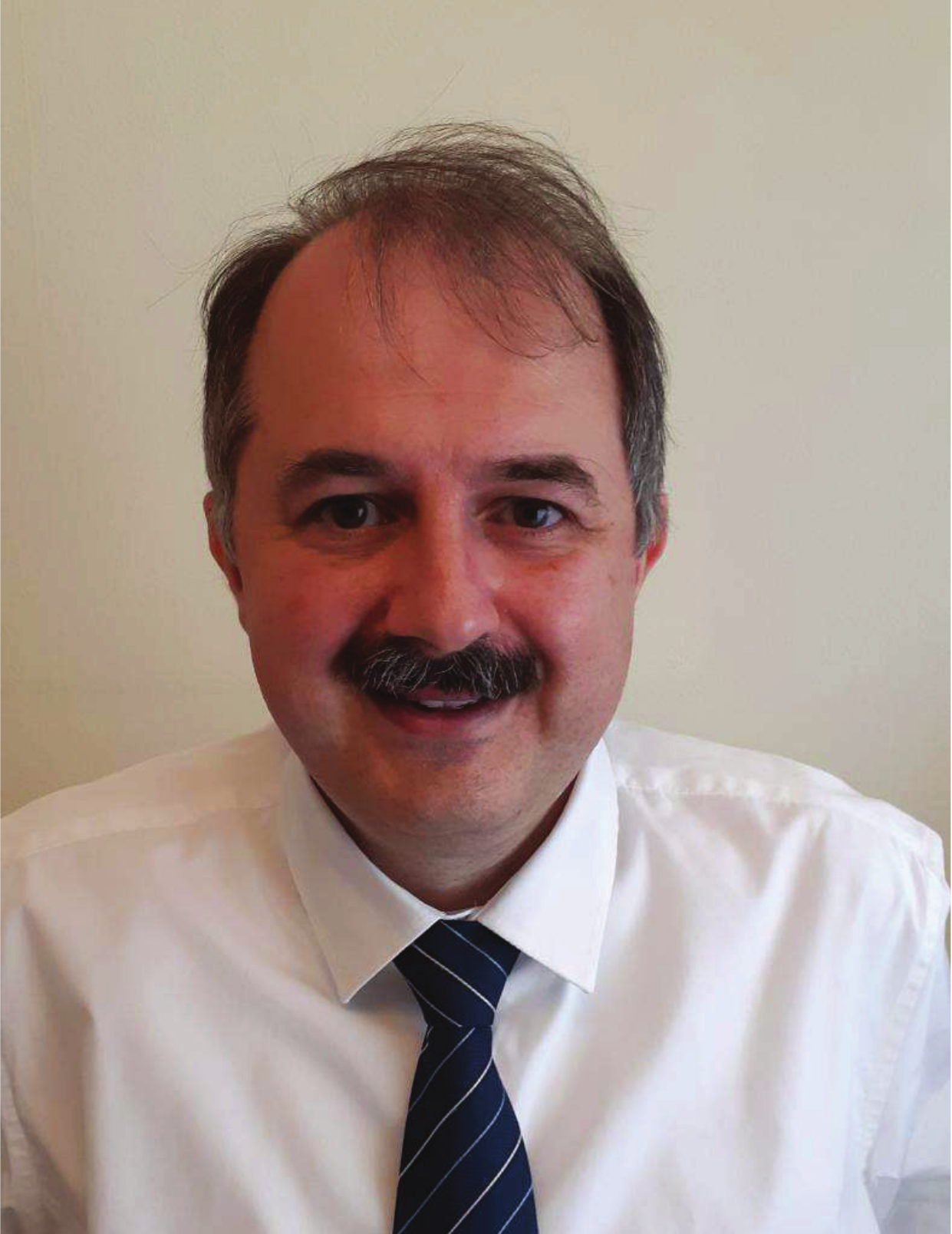}}]{Gabriel-Miro Muntean} [F 22] is a Professor with the School of Electronic Engineering, Dublin City University (DCU), Ireland, and co-Director of DCU Performance Engineering Laboratory. He has published 4 books and over 450 papers in top international journals and conferences. His research interests include rich media delivery quality, performance, and energy-related issues, technology enhanced learning, and other data communications in heterogeneous networks. He is an Associate Editor of the \textsc{IEEE Transactions on Broadcasting}, the Multimedia Communications Area Editor of the \textsc{IEEE Communications Surveys and Tutorials}, and reviewer for important international journals, conferences, and funding agencies. He coordinated the EU project NEWTON and leads the DCU team in the EU project TRACTION.
\end{IEEEbiography}

\vspace{-3em}

\begin{IEEEbiography}[{\includegraphics[width=1in,height=1.2in,clip,keepaspectratio]{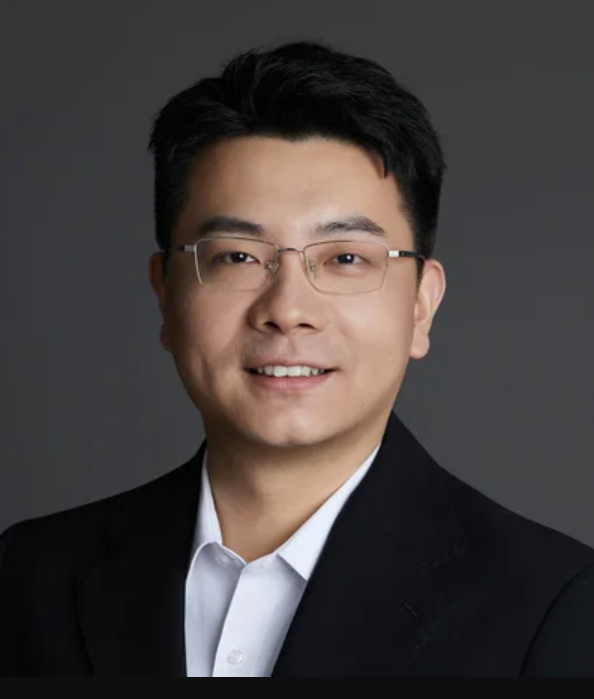}}]{Enmao Diao} was born in Chengdu, Sichuan, China, in 1994. He received the B.S. degree in computer science and electrical engineering from the Georgia Institute of Technology, Atlanta, GA, USA, in 2016, the M.S. degree in electrical engineering from Harvard University, Cambridge, MA, USA, in 2018, and the Ph.D. degree in electrical engineering from Duke University, Durham, NC, USA, in 2023. His research interests include distributed machine learning, efficient machine learning, and signal processing.
\end{IEEEbiography}

\end{document}